%% file: main.tex
\documentclass[conference]{IEEEtran}
\IEEEoverridecommandlockouts
\usepackage{cite}
\usepackage{amsmath,amssymb,amsfonts}
\usepackage{algorithmic}
\usepackage{graphicx}
\usepackage{textcomp}
\usepackage{xcolor}
\def\BibTeX{{\rm B\kern-.05em{\sc i\kern-.025em b}\kern-.08em
    T\kern-.1667em\lower.7ex\hbox{E}\kern-.125emX}}

\usepackage[hyphens]{url} 
\usepackage{graphicx}
\usepackage{caption}
\usepackage{booktabs}
\usepackage{multirow}
\usepackage{algorithm}
\usepackage{algorithmic}
\usepackage{amsmath}
\usepackage{subcaption}
\usepackage{amsthm}
\usepackage{tabularx, amssymb}
\usepackage{url} 
\usepackage{bm}
\usepackage{newfloat}
\usepackage{listings}
\input{macros}

\begin{document}

\title{Model- and Layer-Level Closed-Loop Forecasting with SSM-Based Spiking Neural Networks\\
\thanks{Computational resources were provided by the National Academic Infrastructure for Supercomputing in Sweden (NAISS), funded by the Swedish Research Council. 
S. Karilanova acknowledges the support of Center for Interdisciplinary Mathematics (CIM), Uppsala University. A.~Özçelikkale acknowledges the support from the Swedish Research Council through grant agreement no. 
2024-05194.}
}

\author{\IEEEauthorblockN{Erik Stolt}
\IEEEauthorblockA{\textit{Department of Electrical Engineering} \\
\textit{Uppsala University}\\
Uppsala, Sweden \\
erikstolt02@gmail.com} % or erik.stolt.6955@student.uu.se
\and
\IEEEauthorblockN{Sanja~Karilanova}
\IEEEauthorblockA{\textit{Department of Electrical Engineering} \\
\textit{Uppsala University}\\
Uppsala, Sweden \\
sanja.karilanova@angstrom.uu.se}
\and
\IEEEauthorblockN{Ayça~Özçelikkale}
\IEEEauthorblockA{\textit{Department of Electrical Engineering} \\
\textit{Uppsala University}\\
Uppsala, Sweden \\
ayca.ozcelikkale@angstrom.uu.se}
}

\maketitle

\input{abstract}
\input{introduction}
\input{related_work}

\input{problem_formulation}
\input{methodology}
\input{experiments}
\input{results}

\input{conclusion}

\input{supplementary_materials}
\bibliographystyle{IEEEtran}
\bibliography{references}

\end{document}

%% file: macros.tex
\newcommand{\Abf}{\bm{A}}
\newcommand{\Bbf}{\bm{B}}
\newcommand{\Cbf}{\bm{C}}

\newcommand{\Wbf}{\matr{W}}
\newcommand{\Kbf}{\matr{K}}

\newcommand{\matr}{\boldsymbol}

\newcommand{\SfourDZM}{S4D-ZM \,}
\newcommand{\SfourDCL}{S4D-Model-CL \,}
\newcommand{\SfourDST}{S4D-Layer-CL \,}

\newcommand{\SfourDZMcomma}{S4D-ZM}
\newcommand{\SfourDCLcomma}{S4D-Model-CL}
\newcommand{\SfourDSTcomma}{S4D-Layer-CL}

\newtheorem{rem}{\bf{Remark}}

\theoremstyle{remark} 

\newenvironment{remark}
{\par\noindent \rem \begin{itshape}\noindent}
{\end{itshape} \vspace{3pt}}

%% file: abstract.tex
\begin{abstract}
Deep State-Space Models (Deep SSMs) have shown promising performance and computational benefits in processing of time-series data with complex temporal dependencies. Spiking neural networks (SNNs) constitute a biologically inspired class of deep SSMs, where computing blocks communicate through spikes, discrete single-bit events rather than continuous activation functions. A particularly important task in time-series processing is forecasting, i.e. predicting future values of a sequence from past observations alone. However, how to best perform forecasting with SNNs is an open question, where the standard autoregressive nature of the data processing with SNNs requires values that are not available during the forecasting phase. In this article, we address this challenge. We explore two closed-loop forecasting strategies, in which the network's own generated values are fed back into the network as if they were observed past values: (i) model-level and (ii) layer-level closed-loop forecasting. We evaluate these approaches on the ETT-small datasets, a standard benchmark for time-series forecasting, under both spike-based and continuous activation functions. We compare the proposed closed-loop approaches against state-of-the-art methods such as SpaceTime, as well as the zero-masking baseline in S4. Both of the proposed approaches perform comparably to or better than zero-masking with S4, and in most cases match SpaceTime. These findings highlight the trade-off between continuous and low-bit quantized activations, and demonstrate that the proposed Deep SSM architectures provide a competitive framework for forecasting.
\end{abstract}

% TL;DR
% We ntroduce two closed-loop forecasting strategies (model-level and layer-level) that feed a spiking neural network's own generated values back into the network, evaluated on the ETT-small dataset, it shows competitive perfromence to SoTA benchamark.

%% file: introduction.tex
\section{Introduction}

Accurate \emph{time-series modeling} is central to signal processing and machine learning, with applications across healthcare, finance, and engineering \cite{shumway2000time}. Real-world sequences frequently exhibit seasonality, trends, and regime shifts, which makes efficient and accurate modeling challenging \cite{shumway2000time}.

\emph{Deep State-Space Models} (Deep SSMs) have recently emerged as effective architectures for long-sequence modeling. They cast sequence processing as a linear dynamical system and scale linearly with sequence length, $\mathcal{O}(T)$, in contrast to the quadratic $\mathcal{O}(T^2)$ complexity of standard Transformer attention \cite{gu2022efficientlymodelinglongsequences,zeng2022transformerseffectivetimeseries}. \emph{Spiking neural networks} (SNNs) can be viewed as a biologically inspired subclass of Deep SSMs whose units communicate through spikes, i.e.\ discrete single-bit events, rather than continuous activations. This makes them attractive for low-power and neuromorphic deployment, and recent work reports substantial energy savings over conventional models on time-series benchmarks \cite{lv2024efficienteffectivetimeseriesforecasting}.

A practical and demanding use of these models is \emph{forecasting}: predicting the next $F$ values, i.e.\ the \emph{forecast horizon}, from the preceding $P$ observations, i.e.\ the \emph{context horizon}. Forecasting is difficult for autoregressive SSMs because the recursion that drives the hidden state requires input values that are simply not available beyond the context. A common workaround is to feed zeros in place of the missing inputs (\emph{zero-masking}), but this discards information and tends to degrade rapidly as the horizon grows. An alternative is to close the loop, feeding the network's own predictions back as surrogate inputs; how best to do this in a Deep SSM, and whether it remains effective under a spike-based activation, is not well understood.

This article investigates closed-loop forecasting with Deep SSMs on real-world data, and studies the effect of the activation function by comparing the continuous GELU activation with a spike-based Signed-Spiked (Signed-Spk.) activation. All models are evaluated with standard forecasting metrics on the ETT-small benchmark \cite{haoyietal-informer-2021}. Our main contributions are as follows.

\begin{itemize}
    \item We formulate and compare two closed-loop forecasting strategies for Deep SSMs: \emph{model-level} closed-loop forecasting, in which the model's output is fed back as the next input, and \emph{layer-level} closed-loop forecasting, in which a dedicated decoder generates the next-step inputs through a trainable feedback matrix, following the design of SpaceTime \cite{zhang2023effectivelymodelingtimeseries}.
    \item We provide a systematic, controlled evaluation of both strategies against a zero-masking baseline on the ETT-small datasets, under identical training budgets and pre-processing.
    \item We study these strategies under both a continuous (GELU) and a spike-based (Signed-Spk.) activation, quantifying the accuracy--efficiency trade-off between continuous and low-bit activations.
\end{itemize}

%% file: related_work.tex
\section{Related Work}

\paragraph{Deep state-space models.}
Structured SSMs have become a strong tool for long-sequence modeling. S4 attains strong results on long-range dependency benchmarks using a structured parameterization of the state matrix \cite{gu2022efficientlymodelinglongsequences}. \cite{gu2022parameterizationinitializationdiagonalstate} subsequently introduced \emph{S4D}, which initializes the state matrix $\matr{A}$ as diagonal; S4D matches S4 in many settings while being simpler to implement and interpret, and we adopt it as our backbone. 

\paragraph{Forecasting with Deep SSMs.}
When S4 was introduced, it was evaluated on time-series forecasting by zero-masking the unknown portion of the input sequence \cite{gu2022efficientlymodelinglongsequences}. \cite{zhang2023effectivelymodelingtimeseries} later proposed \textsc{SpaceTime} (ST), an encoder--decoder architecture tailored to forecasting that combines companion-structured SSMs, residual connections, linear layers, and GELU activations. Crucially, ST introduces a \emph{closed-loop} strategy in which a trainable matrix $\matr{K}$ synthesizes future inputs to advance the decoder past the context. Our layer-level strategy is inspired by this mechanism, but we apply the feedback matrix to the non-activated neuron outputs rather than to the hidden state (see Section \ref{sec:methodology}). More broadly, simple linear models have been shown to be surprisingly competitive with Transformers for forecasting \cite{zeng2022transformerseffectivetimeseries}, motivating the study of lightweight SSM-based alternatives.

\paragraph{Spiking neural networks for forecasting.}
SNNs have been explored primarily for the energy efficiency of time-series prediction \cite{KULKARNI20133628, lv2024efficienteffectivetimeseriesforecasting, shibo2025tslif, wu2025spikf}. \cite{KULKARNI20133628} applied a spiking model to short-term electrical load forecasting, and \cite{shibo2025tslif} introduced a temporal-segment spiking neuron for time series. \cite{lv2024efficienteffectivetimeseriesforecasting} evaluated SNNs across diverse forecasting benchmarks and found them substantially more energy-efficient than traditional methods, highlighting their potential for this task. In contrast to these works, we study how the \emph{forecasting recursion} itself should be closed for SSM-based spiking models, and directly compare a spike-based activation against a continuous one within a single controlled setup.

%% file: problem_formulation.tex
\section{Problem Formulation}
\label{sec:Problem Formulation}

\subsection{Preliminaries}
\label{sec:Preliminaries}

\subsection{Discrete-Time State-Space Models}
\label{subsec:Discrete-Time State-Space Models}
A discrete-time SSM with linear state dynamics and nonlinear output function can be written as

\begin{subequations}  \label{eq:ssm}
\begin{align}
    \Vec{x}[t+1] &= \matr{A} \Vec{x}[t] + \matr{B} \Vec{u}[t], \\
    \label{eq:hidden}
    \Vec{y}[t+1] &= f\bigg( \mathfrak{R} \left( \matr{C} \Vec{x}[t+1] \right) \bigg),
\end{align}
\end{subequations}
where $\matr{A} \in \mathbb{C}^{n \times n}$ is a diagonal state-transition matrix, $\matr{B} \in \mathbb{R}^{n \times n_{\text{in}}}$ maps the input into the state, and $\matr{C} \in \mathbb{R}^{n_{\text{out}} \times n}$ maps the state to the output. The input, state, and output vectors have dimensions $\Vec{u}[t] \in \mathbb{R}^{n_{\text{in}}}$, $\Vec{x}[t] \in \mathbb{C}^{n}$, and $\Vec{y}[t] \in \mathbb{R}^{n_{\text{out}}}$, respectively. Here $\mathfrak{R}(\cdot)$ denotes the real part and $f(\cdot)$ is an arbitrary activation function.

\subsubsection{Deep State-Space Models}
\label{sec:Deep SSM}
\begin{figure}
  \centering
  \includegraphics[width=0.35\textwidth]{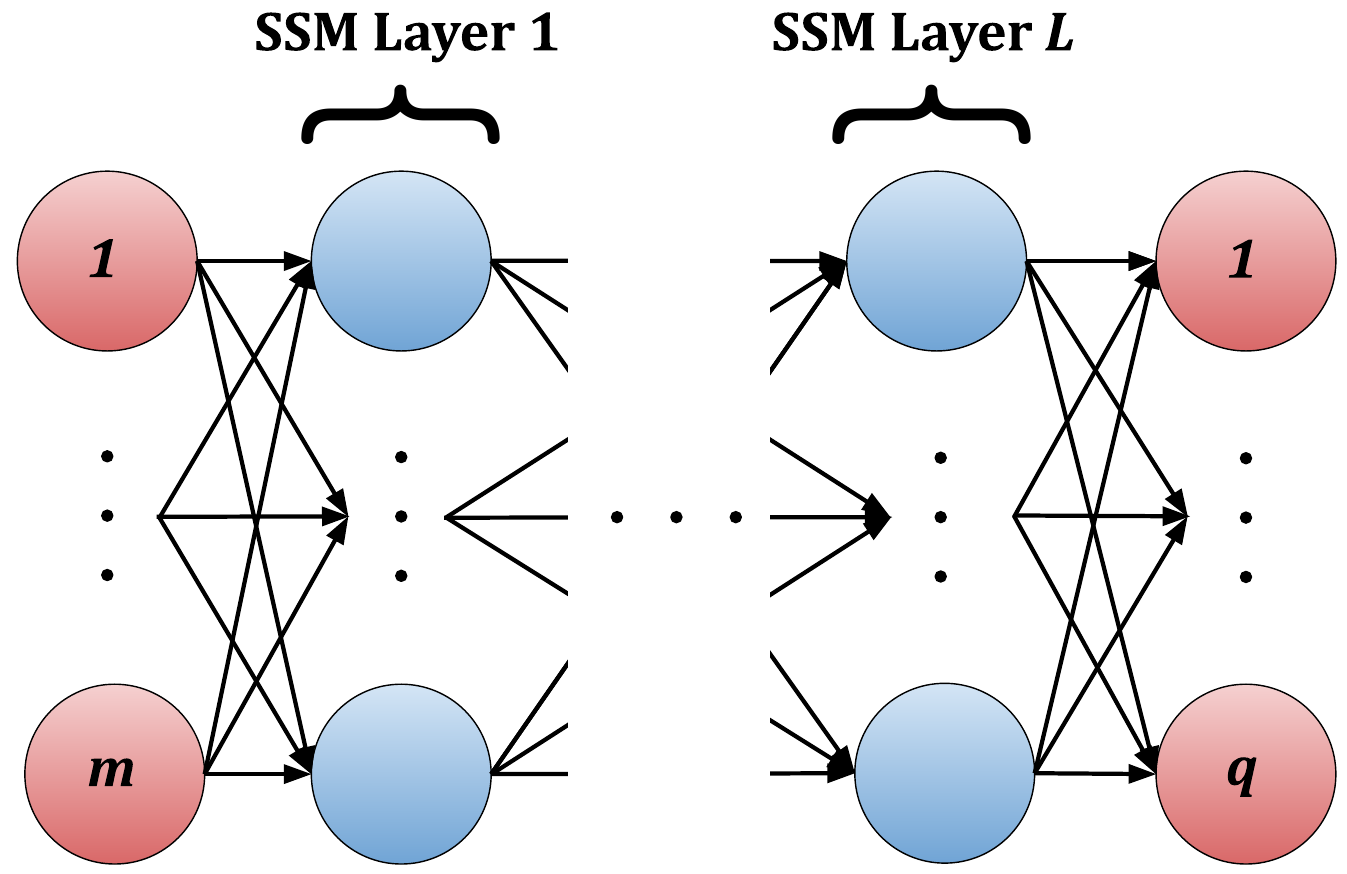}\hfill
  \caption{Deep SSM architecture with an input layer consisting of $m$ input features, followed by $L$ SSM layers, and an output layer consisting of $q$ output features.}
  \label{fig:network}
\end{figure}

A Deep SSM is constructed by embedding SSM-based neurons into a neural-network architecture. The network consists of an input layer of $m$ channels, an output layer of $q$ channels, and $L$ SSM hidden layers. These SSM hidden layers are simply referred to as \emph{SSM layers}. Each SSM layer concatenates $h$ SSM neurons of the form~\eqref{eq:ssm} into an array; in this article all layers use the same number of neurons $h$ and the same state dimension $n$. Stacking $L$ such layers gives the network its depth. Consecutive layers are connected by mixing (weight) matrices $\matr{W} \in \mathbb{R}^{h n_{\text{out}} \times h n_{\text{in}}}$ and batch-normalization layers \cite{ioffe2015batchnormalizationacceleratingdeep}. Figure~\ref{fig:network} illustrates the architecture.

\subsection{Forecasting}
\label{sec:Forecasting}

Let $\Vec{U}[t] \in \mathbb{R}^{m}$ denote the discrete-time input signal with $m$ features for $t = 0, \ldots, T-1$, where $T$ is the sequence length, and let $\Vec{Y}[t] \in \mathbb{R}^{q}$ denote the corresponding $q$-feature output signal. For a forecast reference time $t = \tau$, the context sequence of length $P$ and the target sequence of length $F$ are
\begin{subequations}\label{eqn:u_vector_definition}
\begin{align} 
    \Vec{u}_i &= \left[ \Vec{U}[\tau-P], \ldots, \Vec{U}[\tau-1] \right] \in \mathbb{R}^{P \times m}, \\
    \Vec{y}_i &= \left[ \Vec{Y}[\tau], \ldots, \Vec{Y}[\tau+F-1] \right] \in \mathbb{R}^{F \times q},
\end{align}
\end{subequations}
% respectively, where $i$ indexes the sample.
%
respectively. Here, $i$ indexes the sample, with $i=\tau-t_0+1$ for a rolling forecasting scenario with a start time $t_0$.
The forecasting task is to learn a function $ \Vec{f}_{\theta}: \mathbb{R}^{P \times m} \to \mathbb{R}^{F \times q}$, parameterized by $\theta$, such that the prediction $\Vec{\hat{y}}_i = \Vec{f}_{\theta}(\Vec{u}_i)$ approximates the true future sequence $\Vec{y}_i$. The parameters are learned by minimizing a loss $\mathcal{L}(\cdot)$ between predicted and true targets over the $N$ training samples:
\begin{equation}
    \hat{\theta}
    = \underset{\theta}{\arg\min}\;
    \frac{1}{N} \sum_{i=1}^N
    \mathcal{L}\big( \Vec{f}_{\theta}(\Vec{u}_i), \Vec{y}_i \big).
\end{equation}
Here ``approximate'' means that $\Vec{f}_\theta(\Vec{u}_i) \approx \Vec{y}_i$, with closeness measured by an evaluation metric that may or may not coincide with the training loss. In this article we focus on \emph{univariate} forecasting, i.e.\ $m = q = 1$, where future values are predicted from past observed values (the context). This setting defines the forecasting objective and serves as the basis for the modeling approaches that follow.

\begin{remark}
In the rest of the article, we refer to $\Vec{u}_i$ and $\Vec{y}_i$ of Eq.~\eqref{eqn:u_vector_definition}  or dropping the sample index $i$,   $\Vec{u}$ and $\Vec{y}$  as the input and output, respectively. Accordingly,  $\Vec{u}$  is written in terms of its elements and without reference to $\tau$ as  
$\Vec{u} = \left[ u[0], \ldots, u[P-1] \right]$,  and similarly for $\Vec{y}$. 
\end{remark}

\subsection{Forecasting with Deep State-Space Models}
\label{sec:Forecasting in Deep SSMs}
Consider a single SSM neuron  in the first hidden layer.
At each time step the SSM receives an input $u[t]$ and, through the hidden state $x[t]$ and the parameters $\matr{A}$, $\matr{B}$, $\matr{C}$, produces an output $y[t]$, as in Eq.~\eqref{eq:ssm}. Since the observed input $u[t]$ is available only for $t = 0, \ldots, P-1$, the recursion cannot be advanced beyond the context without an assumption about the unseen future inputs $u[t]$, $t \geq P$. Hence, a network of SSM neurons cannot provide a forecasting output. 
The forecasting strategies in the next section differ precisely in how they supply these missing inputs to the network. 

%% file: methodology.tex
\section{Methodology}
\label{sec:methodology}
We consider three ways of supplying the missing future inputs to a Deep SSM: a zero-masking baseline, and two closed-loop strategies that feed the network's own predictions back as inputs. We describe each in turn.

\subsection{Baseline: Zero-Masking}
\label{subsec:Long-Horizon Forecasting Using Zero-Masking}

The simplest strategy expands the input vector and assumes zero-valued inputs for all future time steps, i.e.\ $u[t] = 0$ for $t \geq P$:
\begin{equation}
    u_{i,\text{mask}}
    = u_{i} \bigoplus 0_F
    \equiv \left[ u_i[0], \ldots, u_i[P-1], 0, \ldots, 0 \right]^{\!\top},
\label{eq:concat}
\end{equation}
where $i$ indexes the sample, $\bigoplus$ is concatenation, $P$ is the context window, $F$ the forecast horizon, and $0_F$ a length-$F$ sequence of zeros. The masked input $u_{i,\text{mask}}$ has length $P+F$, so the corresponding output produced by the SSM also has length $P+F$. During the future portion the hidden state evolves under the \emph{autonomous dynamics} $\Vec{x}[t+1] = \matr{A}\,\Vec{x}[t]$, since the input contribution vanishes. We refer to this strategy as \emph{zero-masking} (ZM); it is the simplest baseline in this work and the strategy used by S4 \cite{gu2022efficientlymodelinglongsequences}.

\paragraph{Loss:}
\label{subsubsec:zm loss function}
For ZM we minimize the mean-squared error (MSE) over the forecast horizon,
\begin{equation}
    \mathcal{L}(\hat{y}, y) \equiv \frac{1}{F} \sum_{t=P}^{P+F-1} \left\| \hat{y}[t] - y[t] \right\|^2 ,
    \label{eq:loss}
\end{equation}
which trains the model to produce accurate predictions at the future time steps.

\subsection{Closed-Loop Forecasting}
\label{subsec:Closed-Loop Forecasting}

\emph{Autoregressive rollout} generates future predictions by feeding each prediction back as the input at the next time step. This section explores two variants of this idea, which differ in \emph{where} the feedback enters the network. In both cases forecasting proceeds in two phases. In the \emph{open-loop} phase the context window of length $P$ is propagated through the network; in the subsequent \emph{closed-loop} phase, starting from the last context step, the network generates its own inputs.

\paragraph{Open-loop phase}
\label{subsubsec:Open-Loop}
The open-loop phase propagates the observed context sequence $u$ through the network. For each $t = 0, \dots, P-1$, the hidden states of all SSM neurons are updated and the corresponding outputs $y[t]$ in Eq.~\eqref{eq:ssm} are produced. Once the context has been consumed, the closed-loop phase begins.
 
\paragraph{Closed-loop phase}
\label{subsubsec:Closed-Loop}
In the closed-loop phase the model generates the input for the next time step. In \emph{model-level} closed-loop forecasting this is simply the model's prediction at the previous step. In \emph{layer-level} closed-loop forecasting, the next-step inputs are instead produced by the decoder through an additional trainable matrix $\matr{K}$. We detail both below.

\subsection{Model-Level Closed-Loop Forecasting}
\label{subsec:Model-Level}
In the model-level variant, the Deep SSM first consumes the $P$-step context to update the hidden state of every SSM neuron. At $t = P$ forecasting begins, and the network output at time $t$ is fed back as the input at time $t+1$. Future predictions are therefore computed as
\begin{equation}
    \hat{y}[t+1] = f_\theta\!\left( \hat{y}[t] \right), \quad t = P, \ldots, P+F-1 .
\end{equation}
This mechanism is entirely dependent on the accuracy of each prediction, since errors are fed back into the model and accumulate over the forecast horizon. The model-level variant is trained with the same loss as ZM, Eq.~\eqref{eq:loss}.

\subsection{Layer-Level Closed-Loop Forecasting}
\label{subsec:Layer-Level}
The layer-level variant is inspired by \textsc{SpaceTime} (ST) \cite{zhang2023effectivelymodelingtimeseries}, which uses an encoder--decoder structure. We assign the first $L-1$ SSM layers to the encoder and the last SSM layer to the decoder. %, with a mixing matrix between consecutive layers.

\paragraph{Encoder}
\label{subsubsec:Encoder}

The encoder follows the standard forward pass of a deep SSM. It consists of $h$ SSM neurons in each of its $L-1$ layers and processes the input sequence $u$ of length $P$ using the recurrence in Eq.~\eqref{eq:ssm} for $t=0,\ldots,P-1$. Between consecutive encoder layers, the neuron outputs are transformed by a mixing matrix $\matr{W}^{(\ell)}$ and bias $\Vec{b}^{(\ell)}$, for $\ell\in{1,\ldots,L-2}$.

\paragraph{Decoder}
\label{subsubsec:Decoder}
The decoder processes the encoded output of the $j$-th encoder neuron, denoted $\bar{u}_j[t] \in \mathbb{R}$ for $j \in \{1, \ldots, h\}$. Unlike the encoder, the decoder appends a trainable matrix $\matr{K}$ that generates the next-step inputs from the current neuron outputs:

\begin{subequations}\label{eq:model}
\begin{align}
    \Vec{x}_j[t+1] &= \matr{A}_j\,\Vec{x}_j[t] + \matr{B}_j\,\bar{u}_j[t],
    \label{eq:dec_state} \\
    z_j[t+1] &= \mathfrak{R}\!\left( \matr{C}_j\,\Vec{x}_j[t+1] \right), \\
    y_j[t+1] &= f\!\left( z_j[t+1] \right), \label{eq:dec_y} \\
    \begin{bmatrix} \hat{\bar{u}}_1[t+1] \\ \vdots \\ \hat{\bar{u}}_h[t+1] \end{bmatrix}
    &= \matr{K} \begin{bmatrix} z_1[t+1] \\ \vdots \\ z_h[t+1] \end{bmatrix}
    \label{eq:dec_K}
\end{align}
\end{subequations}
for $t = 0, \ldots, P+F-1$, with $\hat{\bar{u}}_j[t] \in \mathbb{R}$ and $\matr{K} \in \mathbb{R}^{h \times h}$. Here $\hat{\bar{u}}_j[t]$ is the predicted next-step input to the $j$-th decoder neuron. The matrix $\matr{K}$ adds $h^2$ trainable parameters relative to the ZM and model-level models, i.e.\ one entry per pair of decoder neurons. See Appendix for the number of trainable parameters introduced by the rest of the parameters in the model. 
 
Our decoder differs from the original ST design in Eq.~\eqref{eq:dec_K}: whereas ST applies $\matr{K}$ to the hidden state of each neuron, we apply it to the vector of \emph{non-activated} neuron outputs $z_j$. Since these non-activated outputs are the quantities that directly form the forecast, applying $\matr{K}$ at this point creates a more direct link between the current decoder predictions and the inputs used at the next forecasting step.

\paragraph{Loss.}
\label{subsubsec:loss function}
The layer-level loss combines three MSE terms: one aligning the encoder outputs with the decoder-generated outputs, one supervising predictions during the open-loop phase, and one supervising predictions during the closed-loop phase:
\begin{equation}
\begin{aligned}
    \mathcal{L}(\hat{y}, y, \hat{\bar{u}}, \bar{u}) =\;& \frac{1}{P-1} \sum_{t=1}^{P-1} \left\| \hat{y}[t] - y[t] \right\|^2 \\
    &+ \frac{1}{P-1} \sum_{t=1}^{P-1} \left\| \hat{\bar{u}}[t] - \bar{u}[t] \right\|^2 \\
    &+ \frac{1}{F} \sum_{t=P}^{P+F-1} \left\| \hat{y}[t] - y[t] \right\|^2 .
\end{aligned}
\end{equation}
This loss encourages accurate predictions during both the open- and closed-loop phases while also supervising the model-generated inputs.

\begin{remark}
The difference between model- and layer-level closed-loop forecasting lies in the feedback mechanism. In model-level closed-loop forecasting, predictions are directly fed back into the first layer of the model. In layer-level closed-loop forecasting, however, the feedback mechanism is handled exclusively by the decoder by feeding back $\hat{\bar{u}}$ as inputs to the decoder in the next time step.
\end{remark}

%% file: experiments.tex
\section{Experiments Set-Up}

\subsection{Architecture}
All models use a Deep SSM with three SSM layers of $64$ neurons each and a state dimension of $128$ per neuron, together with linear input and output projections. The state matrix $\matr{A}$ is initialized as S4D-Lin \cite{gu2022parameterizationinitializationdiagonalstate}, $\matr{B}$ is initialized to all ones, and $\matr{C}$ is initialized randomly with unit standard deviation. In layer-level closed-loop forecasting, $\matr{K}$ is initialized using Kaiming uniform initialization. 
The trainable parameter in the models constitute: for each SSM neurons $\Abf^l_j, \Bbf^l_j, \Cbf^l_j$ and when present $\Kbf^l_j$ for $j=1,...h$ neurons and $l=1,...L$ hidden layers; the weights with bias $\Wbf^l$ for $l=1,...L+1$ connecting each layer in the network; and the trainable weight and bias in the normalization layer.
% and subsequently optimized during training. 
The same architecture and initialization are used for every model in this work, so that differences in performance can be attributed to the forecasting strategy and the activation function rather than to capacity. 
Please see Appendix for hyperparameters used and the number of trainable parameters overview.  All models are trained on NVIDIA Tesla T4 GPU with 16GB RAM. 

We compare two activation functions: the continuous GELU and the spike-based Signed-Spiked activation,
\begin{equation}
\operatorname{Signed\text{-}Spiked}(x)
=
\operatorname{sgn}(x)\,
\mathbb{I}\!\left(|x| > \vartheta\right),
\label{eq:signed_spiked}
\end{equation}
where $\vartheta \in \mathbb{R}$ is the firing threshold set to $\vartheta=1$ and $\mathbb{I}(\cdot)$ the indicator function. The Signed-Spiked activation emits a signed single-bit event ($-1$, $0$, or $+1$), yielding low-bit inter-layer communication. As this function is non-differentiable, we train it with a box-car surrogate gradient in the backward pass \cite{neftci2019surrogate, 10.3389/fnins.2022.865897}, while in the forward pass we use the spiking function.

\subsection{Datasets}
\label{subsubsec:subsec:the ETTm1 Dataset}
We evaluate on the univariate Small Electricity Transformer Temperature (ETT-small) benchmark \cite{haoyietal-informer-2021}, which comprises the datasets ETTm1, ETTm2, ETTh1, and ETTh2 (provided as CSV). We omit ETTm2 since it shares many characteristics with ETTm1 and due to its size. These datasets record long time series from two electricity transformers at two stations over roughly $20$ months \cite{haoyietal-informer-2021}. They are widely used benchmarks for long-range dependency (LRD) tasks because they combine long- and short-term periodic patterns with numerous irregular fluctuations. Each dataset is split $0.6/0.2/0.2$ into train/validation/test partitions and normalized using the mean and variance calculated from the training dataset.

\subsection{Evaluation Metrics}
The aim of the experiments is to compare diagonal SSM models under different activation functions and forecasting strategy, and to position them relative to these existing methods on the standardized ETT datasets. Hence for all experiments we report the MSE and mean absolute error (MAE) over the forecast horizon as defined in Section \ref{sec:methodology} for each forecasting strategy. The values are reported as average over five random initializations of the model. Comparisons to the S4 and \textsc{SpaceTime} numbers are quote from \cite{zhang2023effectivelymodelingtimeseries}. 

%% file: results.tex
\section{Results}

\begin{table*}[t]
    \centering
    \setlength{\tabcolsep}{4pt}
    \caption{Univariate time-series forecasting results on three ETT-small datasets. Results are averaged over five random initializations; subscripts denote the standard deviation. Results for S4 and \textsc{SpaceTime} are taken from~\cite{zhang2023effectivelymodelingtimeseries}. \textbf{Bold} denotes the best and \underline{underline} the second-best model per row.}
    \label{tab:results}

    \begin{subtable}{\textwidth}
        \centering
        \caption{Context lengths $P \in \{48, 96, 720\}$ and forecast horizon $F=24$. S4 and \textsc{SpaceTime} results from Table~7 in~\cite{zhang2023effectivelymodelingtimeseries}.}
        \label{tab:F24}
        \begin{tabular}{lll cc cc cc cc cc}
            \toprule
            & & & \multicolumn{2}{c}{\textbf{\SfourDZM}}
                & \multicolumn{2}{c}{\textbf{\SfourDCL}}
                & \multicolumn{2}{c}{\textbf{\SfourDST}}
                & \multicolumn{2}{c}{\textbf{S4}}
                & \multicolumn{2}{c}{\textbf{\textsc{SpaceTime}}} \\
            \cmidrule(lr){4-5} \cmidrule(lr){6-7} \cmidrule(lr){8-9} \cmidrule(lr){10-11} \cmidrule(lr){12-13}
            Activation & Dataset & $P$ & MSE & MAE & MSE & MAE & MSE & MAE & MSE & MAE & MSE & MAE \\
            \midrule
            \multirow{3}{*}{GELU}
            & ETTh1 & 720 & $0.08_{\pm .00}$ & $0.22_{\pm .01}$ & $0.08_{\pm .00}$ & $0.23_{\pm .00}$ & $\underline{0.06}_{\pm .00}$ & $\underline{0.18}_{\pm .01}$ & $0.06$ & $0.19$ & $\bm{0.03}$ & $\bm{0.12}$ \\
            & ETTh2 & 48  & $0.12_{\pm .00}$ & $0.26_{\pm .01}$ & $0.12_{\pm .00}$ & $0.25_{\pm .01}$ & $0.11_{\pm .00}$ & $0.24_{\pm .00}$ & $\underline{0.10}$ & $\underline{0.23}$ & $\bm{0.06}$ & $\bm{0.19}$ \\
            & ETTm1 & 96  & $0.02_{\pm .00}$ & $0.09_{\pm .00}$ & $0.02_{\pm .00}$ & $0.09_{\pm .00}$ & $\underline{0.02}_{\pm .00}$ & $\underline{0.09}_{\pm .00}$ & $0.02$ & $0.12$ & $\bm{0.01}$ & $\bm{0.07}$ \\
            \midrule
            \multirow{3}{*}{Signed-Spk.}
            & ETTh1 & 720 & $\underline{0.13}_{\pm .01}$ & $\underline{0.28}_{\pm .01}$ & $0.13_{\pm .01}$ & $0.29_{\pm .01}$ & $\bm{0.11}_{\pm .01}$ & $\bm{0.26}_{\pm .02}$ & -- & -- & -- & -- \\
            & ETTh2 & 48  & $0.13_{\pm .00}$ & $0.27_{\pm .00}$ & $\underline{0.13}_{\pm .00}$ & $\underline{0.27}_{\pm .00}$ & $\bm{0.13}_{\pm .00}$ & $\bm{0.27}_{\pm .00}$ & -- & -- & -- & -- \\
            & ETTm1 & 96  & $0.02_{\pm .00}$ & $0.11_{\pm .00}$ & $\bm{0.02}_{\pm .00}$ & $\bm{0.10}_{\pm .00}$ & $\underline{0.02}_{\pm .00}$ & $\underline{0.10}_{\pm .00}$ & -- & -- & -- & -- \\
            \bottomrule
        \end{tabular}
    \end{subtable}

    \vspace{1em}

    \begin{subtable}{\textwidth}
        \centering
        \caption{Context length $P=720$ and forecast horizon $F=96$. S4 and \textsc{SpaceTime} results from Table~1 in~\cite{zhang2023effectivelymodelingtimeseries}.}
        \label{tab:F96}
        \begin{tabular}{ll cc cc cc cc cc}
            \toprule
            & & \multicolumn{2}{c}{\textbf{\SfourDZM}}
              & \multicolumn{2}{c}{\textbf{\SfourDCL}}
              & \multicolumn{2}{c}{\textbf{\SfourDST}}
              & \multicolumn{2}{c}{\textbf{S4}}
              & \multicolumn{2}{c}{\textbf{\textsc{SpaceTime}}} \\
            \cmidrule(lr){3-4} \cmidrule(lr){5-6} \cmidrule(lr){7-8} \cmidrule(lr){9-10} \cmidrule(lr){11-12}
            Activation & Dataset & MSE & MAE & MSE & MAE & MSE & MAE & MSE & MAE & MSE & MAE \\
            \midrule
            \multirow{3}{*}{GELU}
            & ETTh1 & $0.17_{\pm .02}$ & $0.33_{\pm .02}$ & $0.19_{\pm .05}$ & $0.35_{\pm .05}$ & $\underline{0.14}_{\pm .02}$ & $\underline{0.30}_{\pm .02}$ & $0.32$ & $0.49$ & $\bm{0.05}$ & $\bm{0.18}$ \\
            & ETTh2 & $0.35_{\pm .04}$ & $0.48_{\pm .03}$ & $0.35_{\pm .03}$ & $0.47_{\pm .03}$ & $\underline{0.31}_{\pm .02}$ & $\underline{0.44}_{\pm .02}$ & $0.38$ & $0.50$ & $\bm{0.12}$ & $\bm{0.27}$ \\
            & ETTm1 & $0.10_{\pm .02}$ & $0.24_{\pm .02}$ & $0.07_{\pm .01}$ & $0.20_{\pm .01}$ & $\underline{0.06}_{\pm .01}$ & $\underline{0.18}_{\pm .02}$ & $0.65$ & $0.73$ & $\bm{0.03}$ & $\bm{0.12}$ \\
            \midrule
            \multirow{3}{*}{Signed-Spk.}
            & ETTh1 & $\bm{0.18}_{\pm .02}$ & $\bm{0.34}_{\pm .02}$ & $0.20_{\pm .02}$ & $0.36_{\pm .02}$ & $\underline{0.20}_{\pm .03}$ & $\underline{0.35}_{\pm .03}$ & -- & -- & -- & -- \\
            & ETTh2 & $\underline{0.30}_{\pm .02}$ & $0.44_{\pm .02}$ & $\bm{0.30}_{\pm .02}$ & $\bm{0.44}_{\pm .02}$ & $\underline{0.30}_{\pm .04}$ & $\underline{0.44}_{\pm .03}$ & -- & -- & -- & -- \\
            & ETTm1 & $\underline{0.09}_{\pm .00}$ & $\underline{0.23}_{\pm .01}$ & $0.10_{\pm .00}$ & $0.24_{\pm .01}$ & $\bm{0.08}_{\pm .01}$ & $\bm{0.21}_{\pm .02}$ & -- & -- & -- & -- \\
            \bottomrule
        \end{tabular}
    \end{subtable}
\end{table*}

\subsection{Forecasting Performance}

% Intro paragraph
We report three models, \SfourDZMcomma, \SfourDCLcomma, and \SfourDSTcomma, where the suffix denotes the forecasting strategy (zero-masking, model-level closed-loop, and layer-level closed-loop) applied to the same S4D backbone. We also include the S4 \cite{gu2022efficientlymodelinglongsequences} and \textsc{SpaceTime} \cite{zhang2023effectivelymodelingtimeseries} numbers as external references. Table~\ref{tab:F24} reports the short-horizon setting ($F=24$) and Table~\ref{tab:F96} the long-horizon setting ($F=96$).

% Spacetime better
\paragraph{\textsc{SpaceTime} as an upper reference.}
Across every dataset and horizon, \textsc{SpaceTime} attains the lowest errors (e.g. MSE $0.026$ on ETTm1 at $F=96$, Table~\ref{tab:F96}). This is expected: \textsc{SpaceTime} adds a multi-layer perceptron on top of the SSMs, uses a companion (rather than diagonal) state matrix, and has more than double the trainable parameters of our models \cite{zhang2023effectivelymodelingtimeseries}. We therefore treat it as a well-resourced upper reference and focus our analysis on the four parameter-comparable, diagonal-SSM models: \SfourDZMcomma, \SfourDCLcomma, \SfourDSTcomma, and S4.

\paragraph{Short horizon ($F=24$).}
Among our three models with GELU (Table~\ref{tab:F24}), \SfourDST is the strongest of our models, and it also beats S4 on two of three datasets: ETTh1 (MSE $0.057$ vs. $0.061$) and ETTm1 (MSE $0.016$ vs. $0.024$). On ETTh2, however, S4 is the best of the four (MSE $0.095$), with \SfourDST closest among our models (MSE $0.112$). The zero-masking approaches (\SfourDZM and S4) remain competitive at the short horizon. Comparing activations, GELU yields uniformly lower errors than Signed-Spk. at $F=24$ (e.g. on ETTh1, \SfourDST drops from $0.108$ to $0.057$). Within the Signed-Spk.\ block, \SfourDST is best on ETTh1 and ETTh2, while \SfourDCL is marginally best on ETTm1 (MSE $0.021$).

\paragraph{Long horizon ($F=96$).}
The picture changes markedly at $F=96$ (Table~\ref{tab:F96}). All of our three models with GELU outperform S4 by a large margin on every dataset. The gap is most striking on ETTm1, where S4 has MSE $0.651$ while \SfourDST reaches $0.060$, and is also clear on ETTh1 ($0.316$ vs. $0.144$). 
As at $F=24$, \SfourDST is the best of our GELU models on all three datasets. For activations, GELU again leads on ETTh1 and ETTm1, but on ETTh2 the spike-based activation is actually stronger, with all three Signed-Spk.\ models (MSE $\approx 0.302$--$0.304$) below their GELU counterparts (MSE $\ge 0.308$). No single strategy dominates across the Signed-Spk.\ results at this horizon: \SfourDZM is best on ETTh1, \SfourDCL on ETTh2, and \SfourDST on ETTm1, indicating that the best choice depends on the dataset.

\begin{figure*}[t]
    \centering
\begin{subfigure}[b]{0.32\textwidth}
    \centering
    \includegraphics[width=0.99\linewidth]{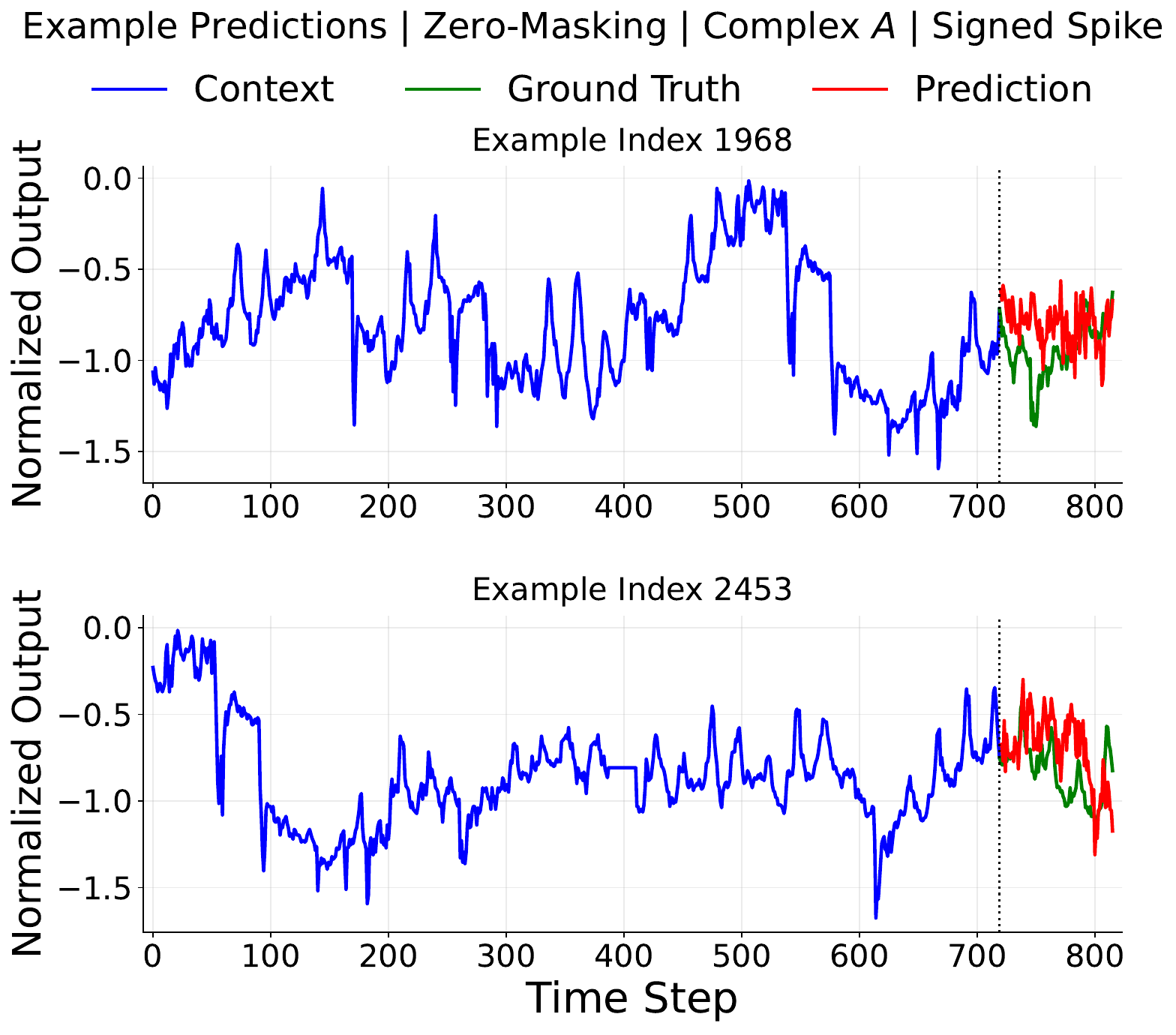}
    \caption{\SfourDZM on ETTh1.}
    \label{fig:ETT1_ZM_SS_0}
\end{subfigure}
\hfill
\begin{subfigure}[b]{0.32\textwidth}
    \centering
    \includegraphics[width=0.99\linewidth]{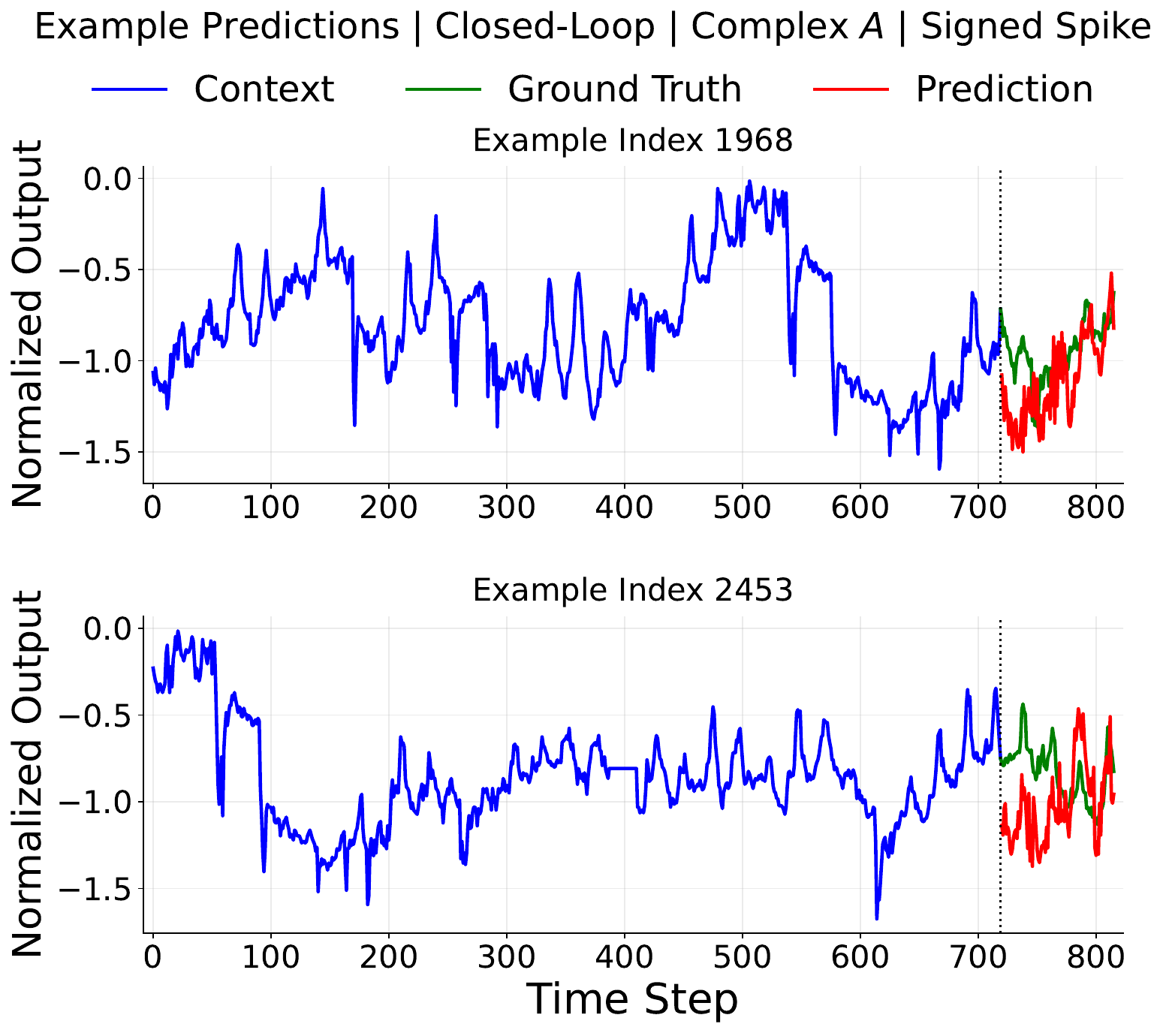}
    \caption{\SfourDCL on ETTh1.}
    \label{fig:ETT1_CL_SS_0}
\end{subfigure}
\hfill
\begin{subfigure}[b]{0.32\textwidth}
    \centering
    \includegraphics[width=0.99\linewidth]{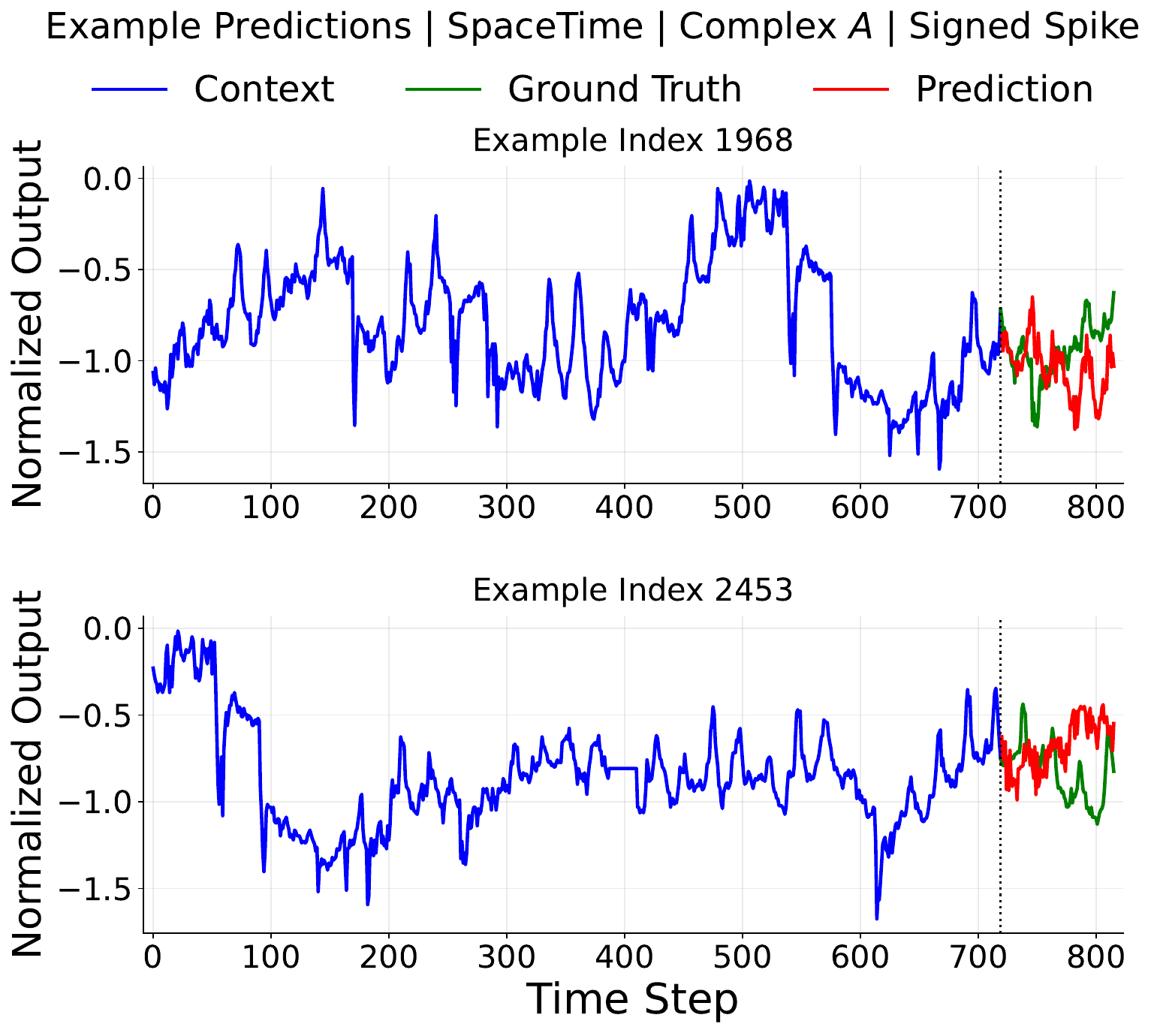}
    \caption{\SfourDST on ETTh1.}
    \label{fig:ETT1_ST_SS_0}
\end{subfigure}
\vspace{0.1cm}
\begin{subfigure}[b]{0.32\textwidth}
    \centering
    \includegraphics[width=0.99\linewidth]{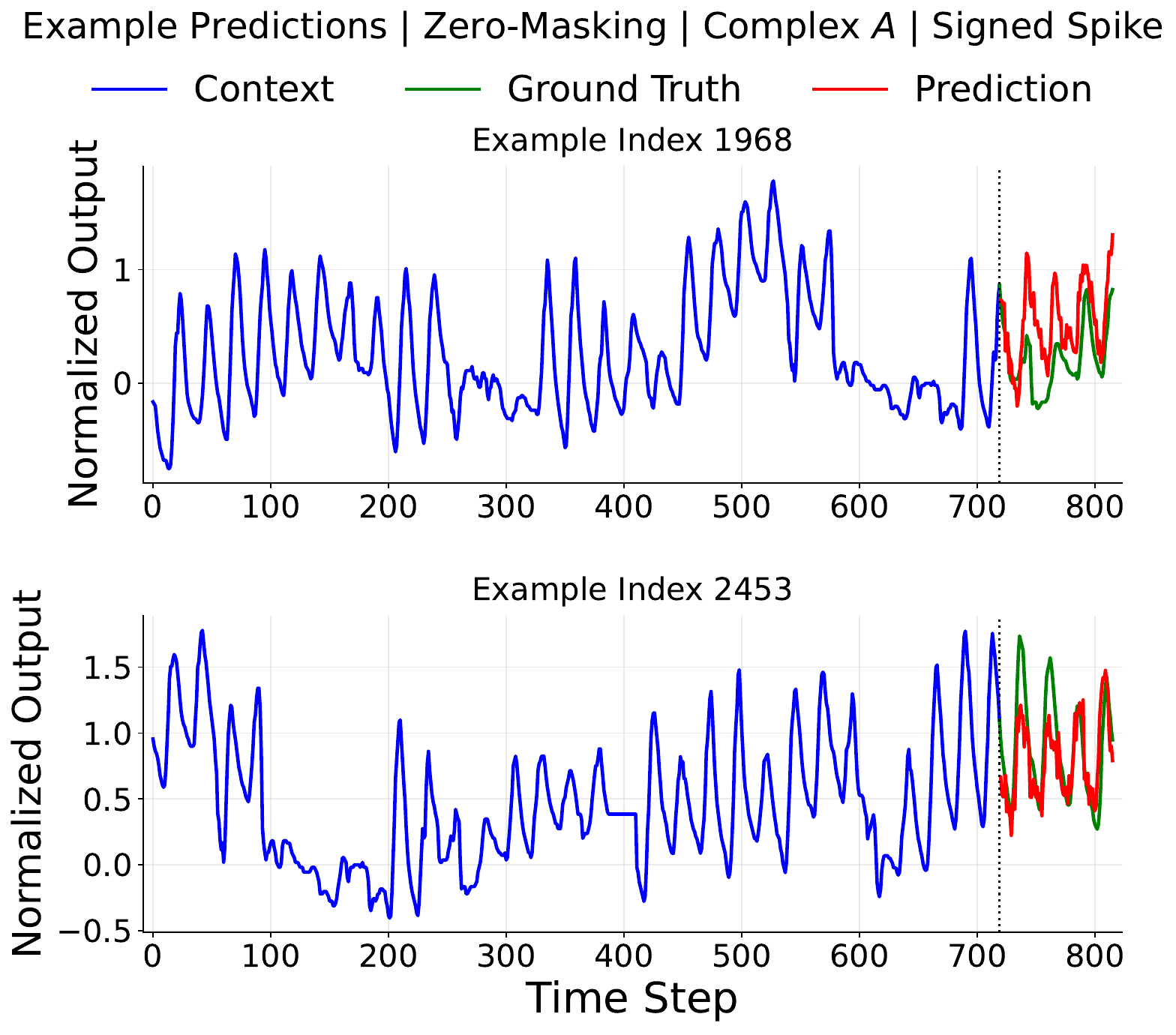}
    \caption{\SfourDZM on ETTh2.}
    \label{fig:ETT2_ZM_SS_0}
\end{subfigure}
\hfill
\begin{subfigure}[b]{0.32\textwidth}
    \centering
    \includegraphics[width=0.99\linewidth]{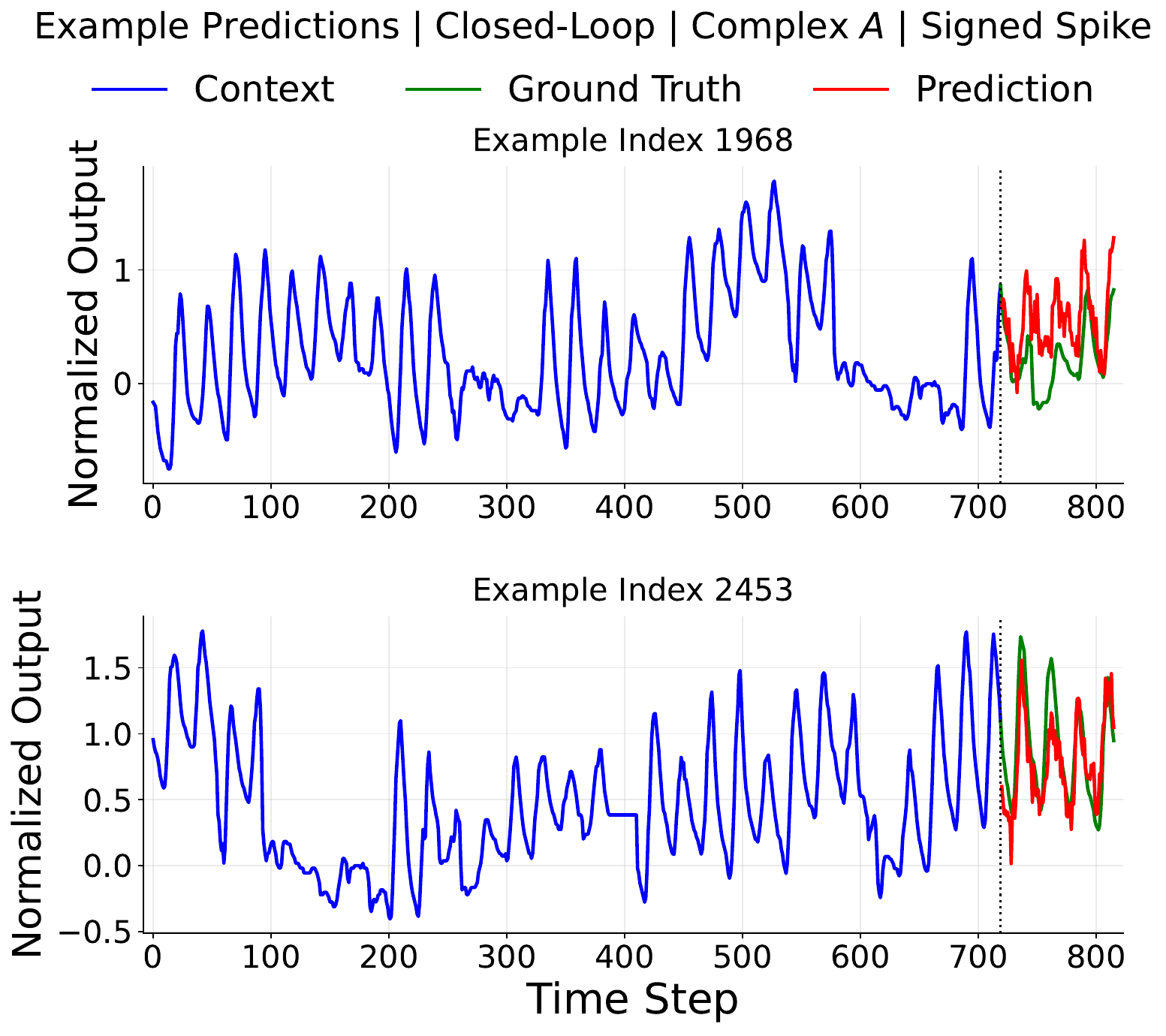}
    \caption{\SfourDCL on ETTh2.}
    \label{fig:ETT2_CL_SS_0}
\end{subfigure}
\hfill
\begin{subfigure}[b]{0.32\textwidth}
    \centering
    \includegraphics[width=0.99\linewidth]{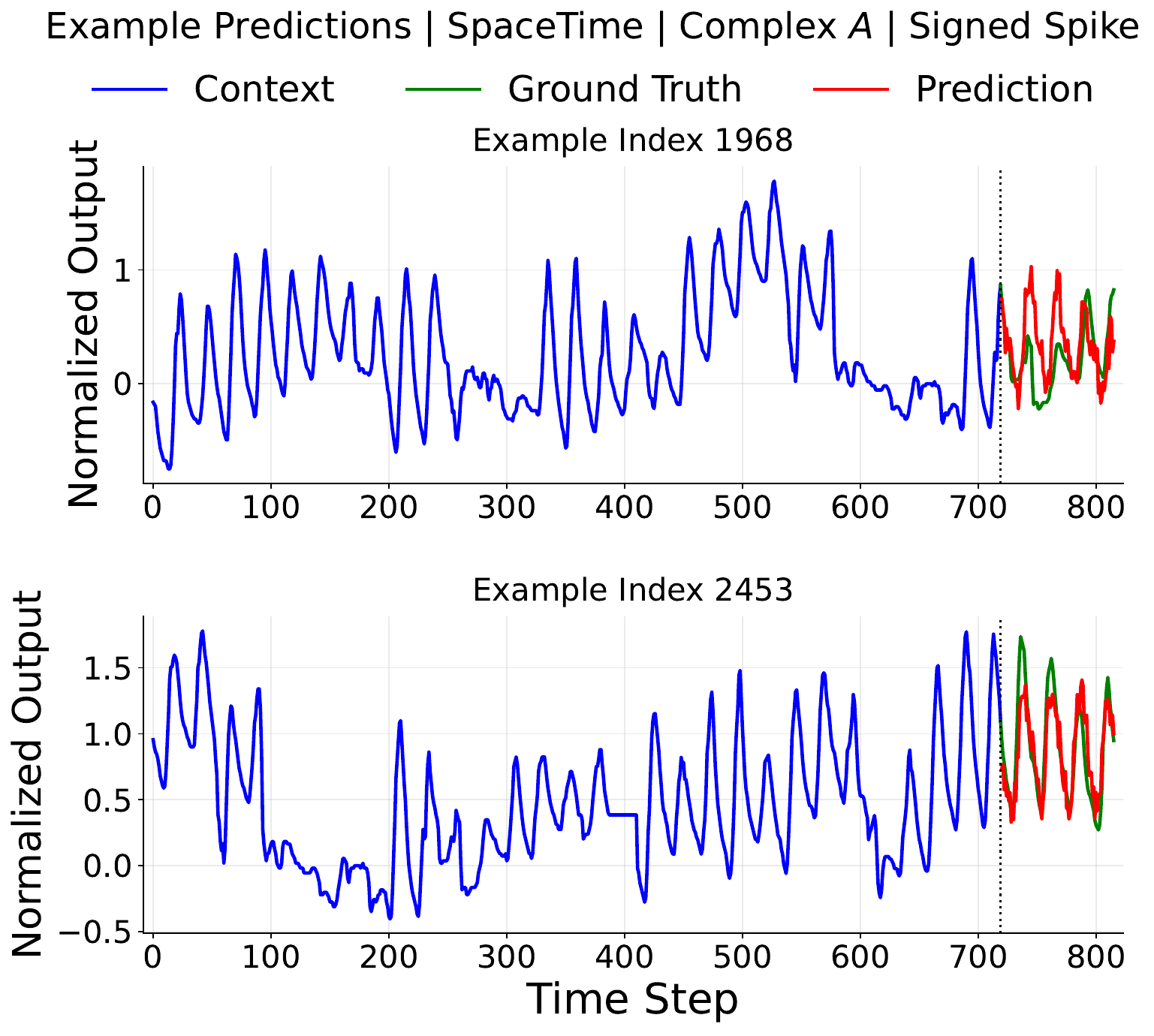}
    \caption{\SfourDST on ETTh2.}
    \label{fig:ETT2_ST_SS_0}
\end{subfigure}
\vspace{0.1cm}
\begin{subfigure}{0.32\textwidth}
    \centering
    \includegraphics[width=0.99\linewidth]{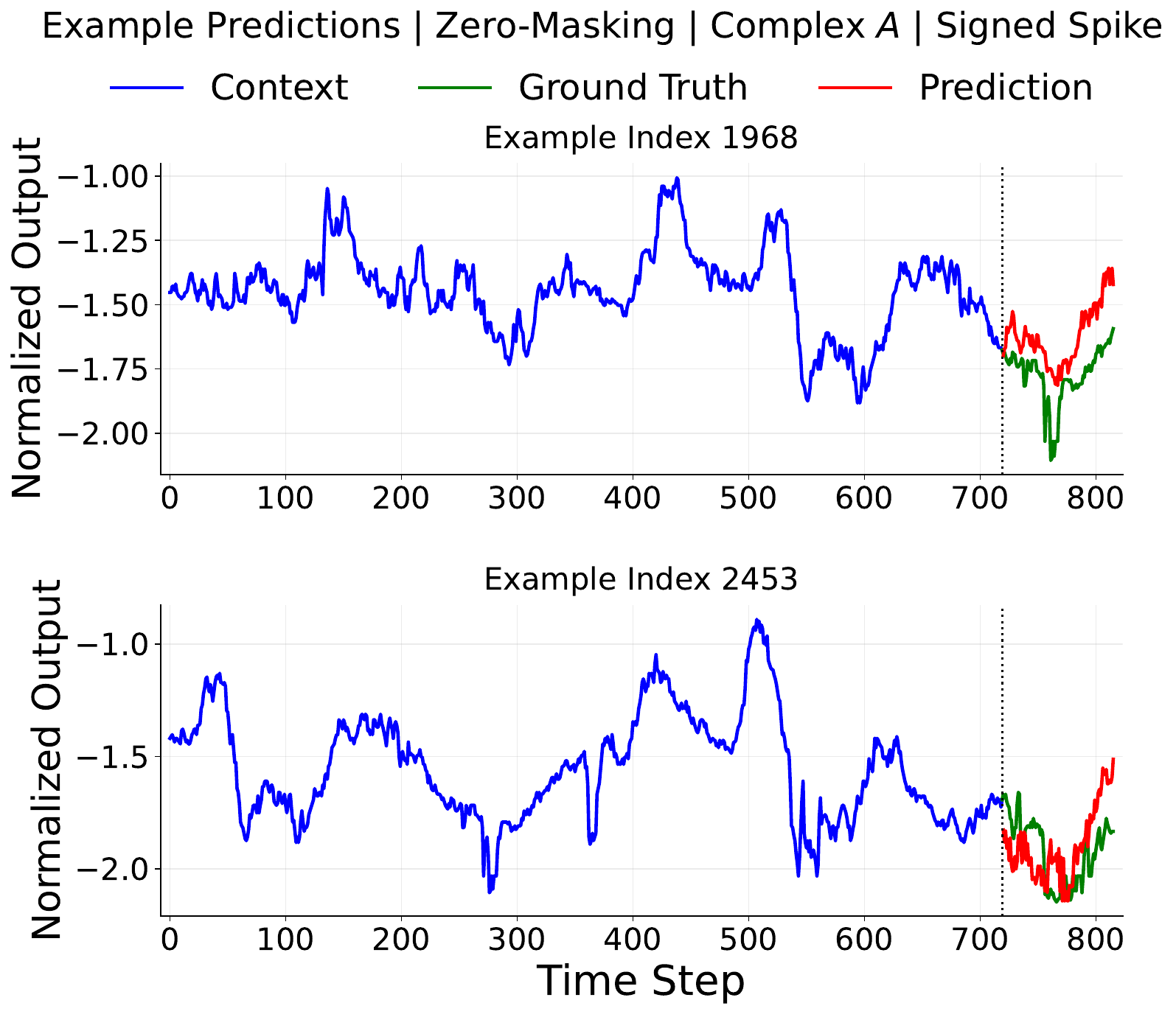}
    \caption{\SfourDZM on ETTm1.}
    \label{fig:ETT3_ZM_SS_0}
\end{subfigure}
\hfill
\begin{subfigure}{0.32\textwidth}
    \centering
    \includegraphics[width=0.99\linewidth]{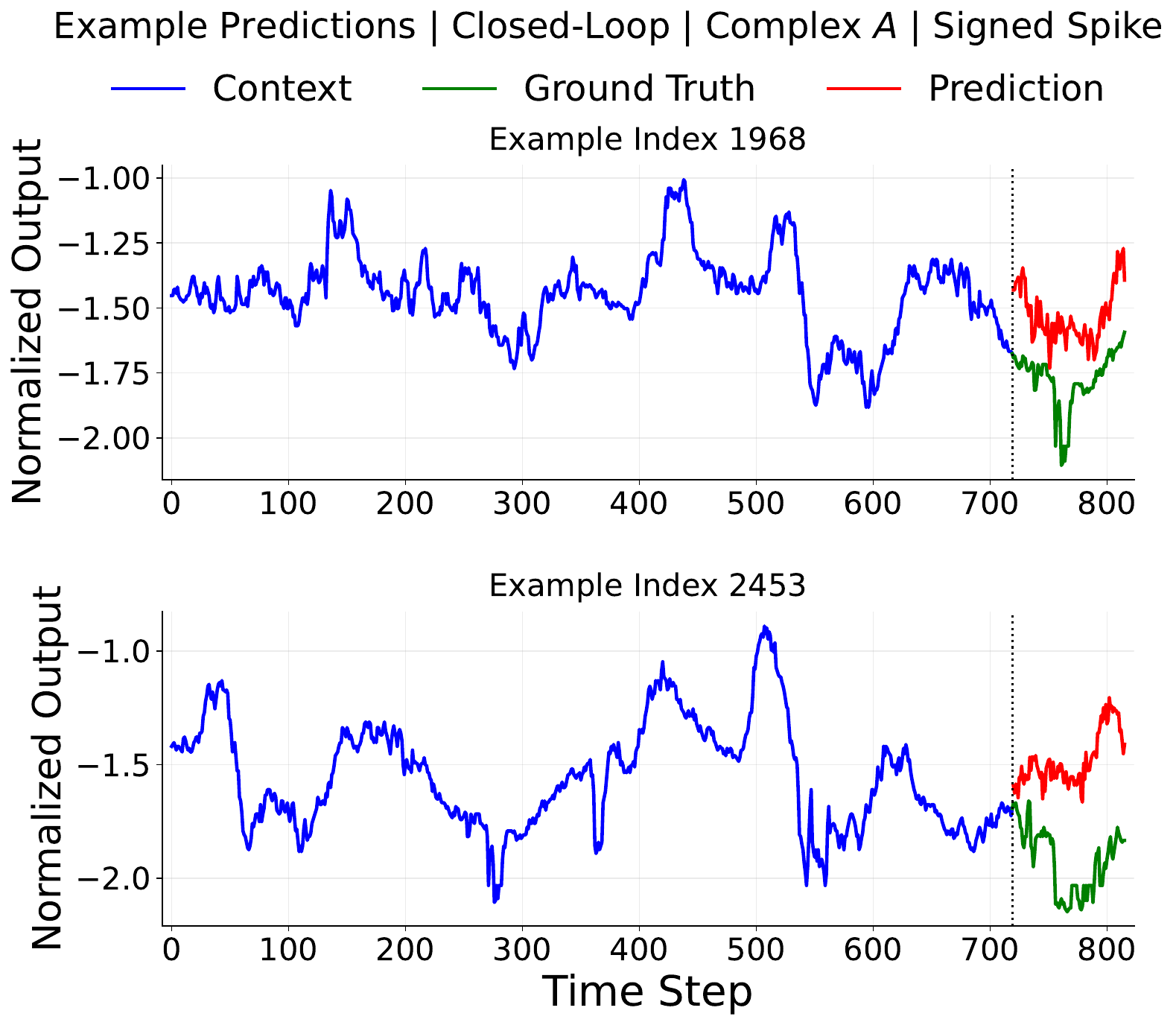}
    \caption{\SfourDCL on ETTm1.}
    \label{fig:ETT3_CL_SS_0}
\end{subfigure}
\hfill
\begin{subfigure}{0.32\textwidth}
    \centering
    \includegraphics[width=0.99\linewidth]{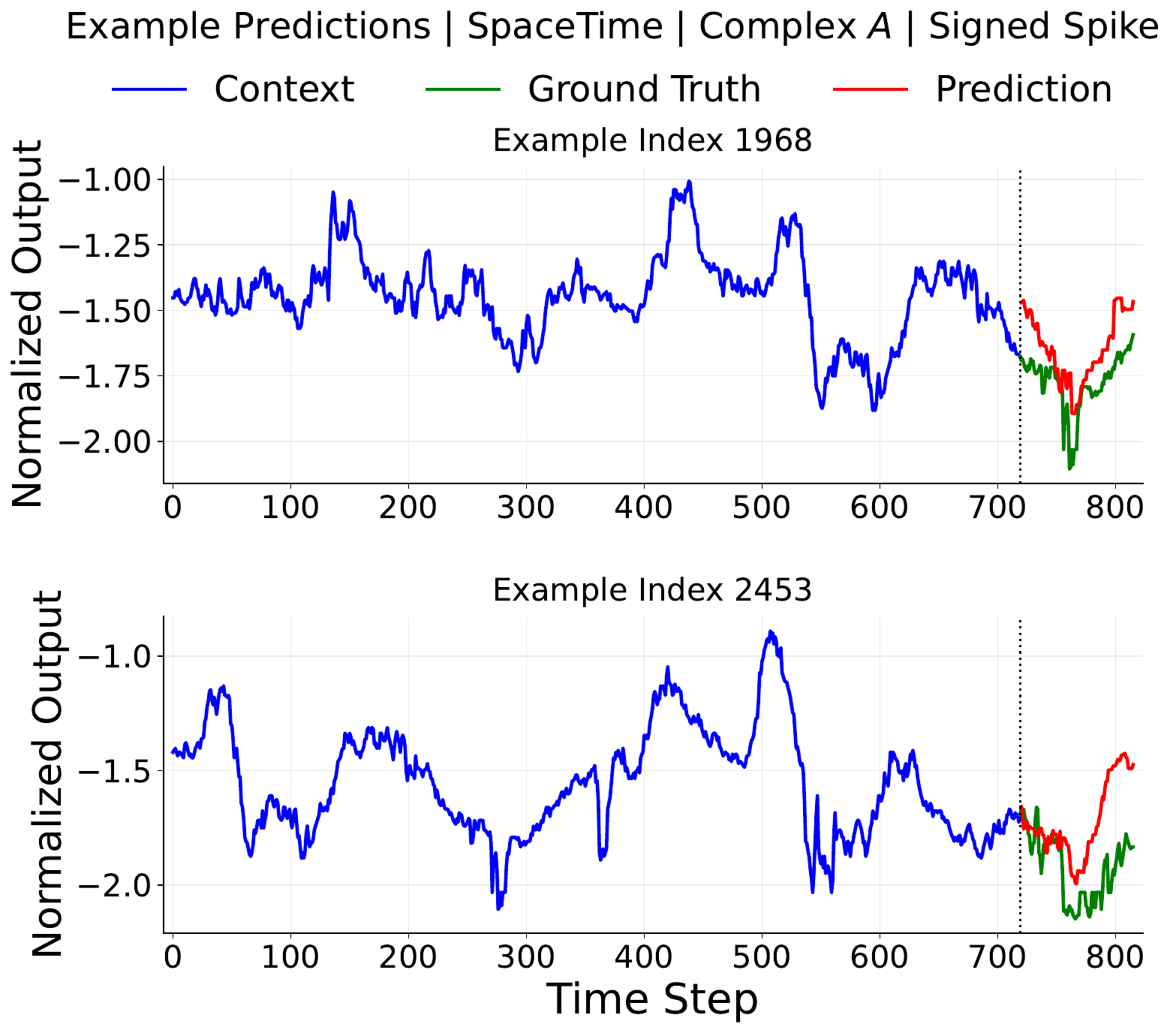}
    \caption{\SfourDST on ETTm1.}
    \label{fig:ETT3_ST_SS_0}
\end{subfigure}
    \caption{Example forecasting samples for all strategies on the ETT-small datasets with context length $P=720$, forecast horizon $F=96$, and  with \textbf{Signed-Spk. activation}.}
    \label{fig:ss}
\end{figure*}

% viusual inspection
\paragraph{Qualitative inspection.}
Figure~\ref{fig:ss} show forecasting examples for $P=720$, and $F=96$ using Signed-Spk. activation. The figure showing the same forecasting examples with GELU activation can be found in Appendix. We can see visually that in many cases the forecasting predictions follow the general pattern of the ground truth.
% In particular on ETTh2 \SfourDST follows the target trajectory closely (Fig.~\ref{fig:ss}).
% \SfourDCL and \SfourDZM also produce faithful rollouts on this dataset. 
The spike-based rollouts in Fig.~\ref{fig:ss} are visibly more quantized than the GELU rollouts, yet remain as aligned with the target.

\begin{figure}
    \centering
    \includegraphics[width=0.99\linewidth]{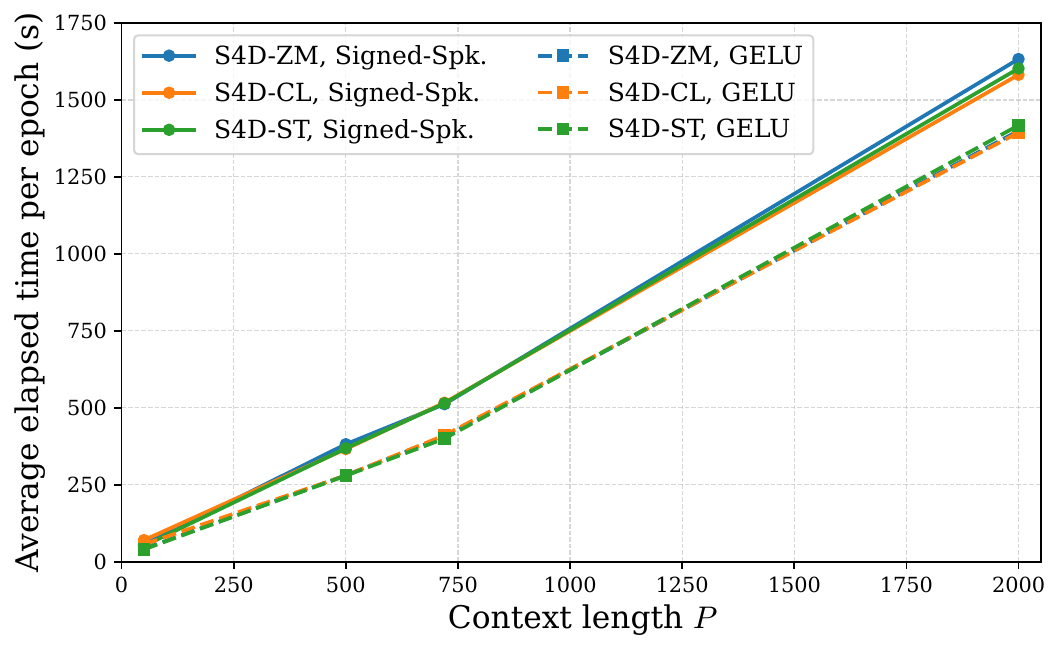}
    \caption{Average elapsed time in seconds per epoch when training on the ETTh1 dataset. All models scale approximately linear with context length.}
    \label{fig:load}
\end{figure}

% Signed-spiked

% --------------------------------------------------------------------
\subsection{Training Time vs. Context Horizon}

We also study empirical training-time scaling with the context length (Fig.~\ref{fig:load}). All models scale approximately linearly with context length, in line with the $\mathcal{O}(T)$ complexity of the SSM recurrence. A small vertical offset separates the GELU and Signed-Spk.\ curves; this might be attributed to our custom implementation of the Signed-Spk.\ activation, whose forward pass and surrogate-gradient backward pass add overhead. However the Signed-Spk. activation is compatible with neuromorphic hardware which at inference can reduces the energy consumption significantly \cite{davies_advancing_2021}.

%% file: conclusion.tex
\section{Conclusion}

We studied how to forecast with SSM-based Deep SSMs by comparing three strategies for supplying the missing future inputs, zero-masking, model-level closed-loop, and layer-level closed-loop, on the same S4D backbone, and under both a continuous (GELU) and a spike-based (Signed-Spk.) activation. On the ETT-small benchmark, the proposed closed-loop models are competitive while scaling approximately linearly with context length. At the short horizon ($F=24$), the layer-level model is generally the strongest of our models and beats S4 on two of three datasets, though zero-masking remains competitive there. At the long horizon ($F=96$), all three of our strategies improve substantially over S4, indicating that closing the loop matters more as the horizon grows. The spike-based activation trails the continuous one in most settings but is competitive at longer horizons and is the stronger choice on ETTh2, illustrating a favorable accuracy--efficiency trade-off for low-bit activations. As expected, the more heavily parameterized \textsc{SpaceTime} remains the best-performing model overall.
 
\paragraph{Limitations and future work.}
Our study is confined to univariate ETT-small forecasting and to two horizons, and our comparisons to S4 and \textsc{SpaceTime} inherit differences in training budget and implementation. Natural next steps are to extend the evaluation to multivariate and longer-horizon settings and additional benchmarks, and to estimate the energy consumption of the spike-based models on neuromorphic hardware.

%% file: supplementary_materials.tex
\section{Appendix}

Table \ref{tab:settings} shows the hyperparameters used in the trained models for which we present the results in the main text. We use hyperparameters motivated by related works.

\begin{table}
\centering
\caption{Model parameters used in the univariate forecasting of ETT-small.}
\small
\begin{tabular}{ll}
\toprule
\textbf{Description} & \textbf{Parameter Value} \\
\midrule    
    Input dimension per SSM neuron & $n_{\text{in}} = 1$ \\
    Output dimension per SSM neuron & $n_{\text{out}}=1$ \\
    Number of stacked SSM neurons & $h=64$ \\
    Number of SSM layers & $L=3$ \\
    State dimension per neuron & $n=128$ \\
    Dropout rate & $0.0$ \\
    Batch size & $B = 64$ \\
    BatchNorm & False \\
    SSM learning rate & $\text{lr}_\text{SSM} = 10^{-2}$ \\
    Other learning rate & $\text{lr}_\text{Other} = 10^{-2}$ \\
    Lr scheduler & Cosine \\
    Minimum lr & $10^{-5}$ \\
    SSM weight decay rate & $\omega_{\text{SSM}} = 10^{-4}$ \\
    Other parameters weight decay rate & $\omega_{\text{Other}} = 10^{-5}$ \\
    Optimizer & AdamW \\
    Number of epochs & $\text{Epochs} = 10$ \\
    Seeds & $0,1,2,3,4$ \\
    SSM output activation function & $f = \text{GELU} / \text{Signed-Spk.}$ \\
    Early stopping & False \\
    Loss & MSE \\
    Context length & $P = \{48, 96, 720\}$ \\
    Forecast horizon & $F \in \{24, 96\}$ \\
    $\text{dt}_\text{min}$ & $10^{-3}$ \\
    $\text{dt}_\text{max}$ & $10^{-1}$ \\
    Target variable & OT \\
    Train/val/test split & $0.6 / 0.2 / 0.2$ \\
\bottomrule
\label{tab:settings}
\end{tabular}
\end{table}

In Table~\ref{tab:notation}, $\ell=0$ corresponds to the input layer and $\ell= L+1$ to the output layer.

\begin{table}
\centering
\caption{Notation used in the model architecture and in network layer $\ell$.}
    % \small
    % \setlength{\tabcolsep}{8pt}  
    \renewcommand{\arraystretch}{1.15}
    %\begin{tabular}{ll}
    \begin{tabular}{l p{5cm}}
    \toprule
    \textbf{Parameter} & \textbf{Description} \\
    \midrule
    $m$ & Number of input features in the input layer \\
    $q$ & Number of output features in the output layer \\
    $n_{\text{in}}$ & Input dimension per SSM neuron  \\
    $n_{\text{out}}$ & Output dimension per SSM neuron  \\
    $h$ & Number of parallel SSM neurons \\
    $L$ & Number of SSM layers \\
    $n$ & State dimension per neuron \\
    $d_{\text{in}} \equiv h n_{\text{in}}$ & Total input dimension to an SSM layer \\
    $d_{\text{out}} \equiv h n_{\text{out}}$ & Total output dimension of an SSM layer \\
    $\ell$ & Layer index $\ell \in \{0, \ldots, L+1 \}$ \\
    $\matr{W}^{(\ell)}$ & Mixing matrix between layer $\ell$ and $\ell + 1$ \\
    $\Vec{b}^{(\ell)}$ & Bias vector between layer $\ell$ and $\ell + 1$ \\
    \bottomrule
 \label{tab:notation}
\end{tabular}
\end{table}

The trainable parameters of the Deep SSM are $\matr{A}, \matr{B}, \matr{C}, \matr{W}^{(\ell)}$, and $\matr{b}^{(\ell)}$ for $\ell = 0, \ldots, L$. For $\matr{W}^{(\ell)}$ and $\matr{b}^{(\ell)}$, the number of trainable parameters equals the number of entries in the corresponding matrix and vector. For $\matr{A}, \matr{B}$ and $\matr{C}$, the number of independent trainable parameters is reduced. In particular, due to the conjugate symmetry discussed by \cite{gu2022parameterizationinitializationdiagonalstate}, only half of the entries are independently parameterized.

The number of trainable parameters is summarized in Table~\ref{tab:trainable}.

\begin{table}
\centering
\small
\caption{Number of trainable parameters for the components of the Deep SSM. $\matr{W}^{(\ell)}$ denotes the mixing matrix between layer $\ell$ and $\ell+1$ with the corresponding bias vector $\Vec{b}^{(\ell)}$.}
\renewcommand{\arraystretch}{1.3}
\begin{tabular}{lc}
    \toprule
    \textbf{Parameter} & \textbf{Number of Trainable Parameters} \\
    \midrule
    $\matr{W}^{(\ell)}  \in \mathbb{R}^{d_{\text{in}} \times d_{\text{out}}}$ & $d_{\text{in}} d_{\text{out}}$ \\
    $\Vec{b}^{(\ell)} \in \mathbb{R}^{d_{\text{in}}}$ & $d_{\text{in}}$ \\
    $\matr{A} \in \mathbb{R}^n$ (diag) & $h n \, / \, 2$ \\
    $\matr{A} \in \mathbb{C}^n$ (diag) & $h n$ \\
    $\matr{B} \in \mathbb{R}^{n \times n_{\text{in}}}$ & $n_{\text{in}} n \, / \, 2$ \\
    $\matr{C} \in \mathbb{R}^{n_{\text{out}} \times n}$ & $n_{\text{out}} n \, / \, 2$ \\
    \bottomrule
\label{tab:trainable}
\end{tabular}
\end{table}

\begin{figure*}[t]
    \centering
\begin{subfigure}[b]{0.32\textwidth}
    \centering
    \includegraphics[width=0.99\linewidth]{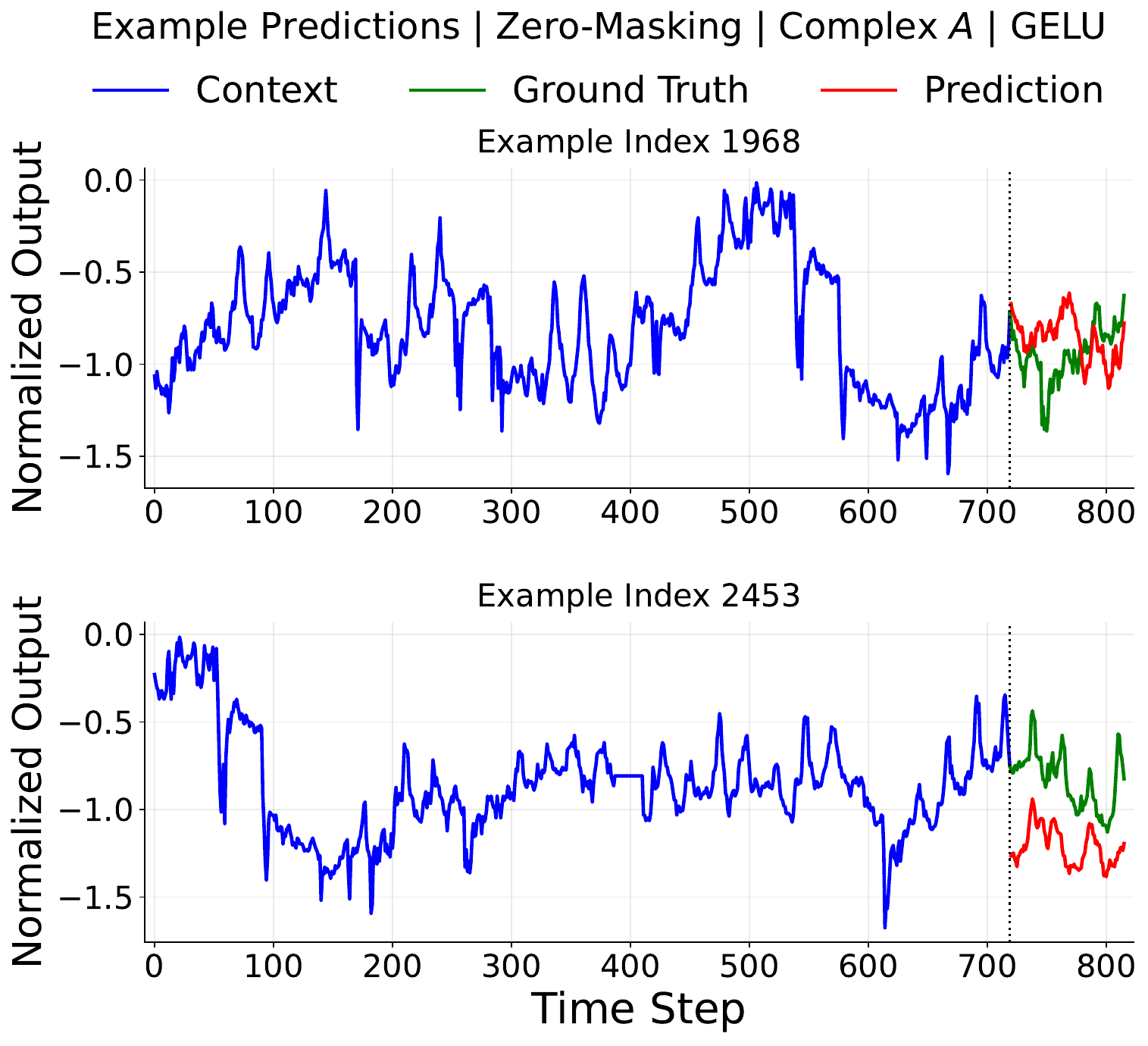}
    \caption{\SfourDZM on ETTh1.}
    \label{fig:ETT1_ZM_G_0}
\end{subfigure}
\hfill
\begin{subfigure}[b]{0.32\textwidth}
    \centering
    \includegraphics[width=0.99\linewidth]{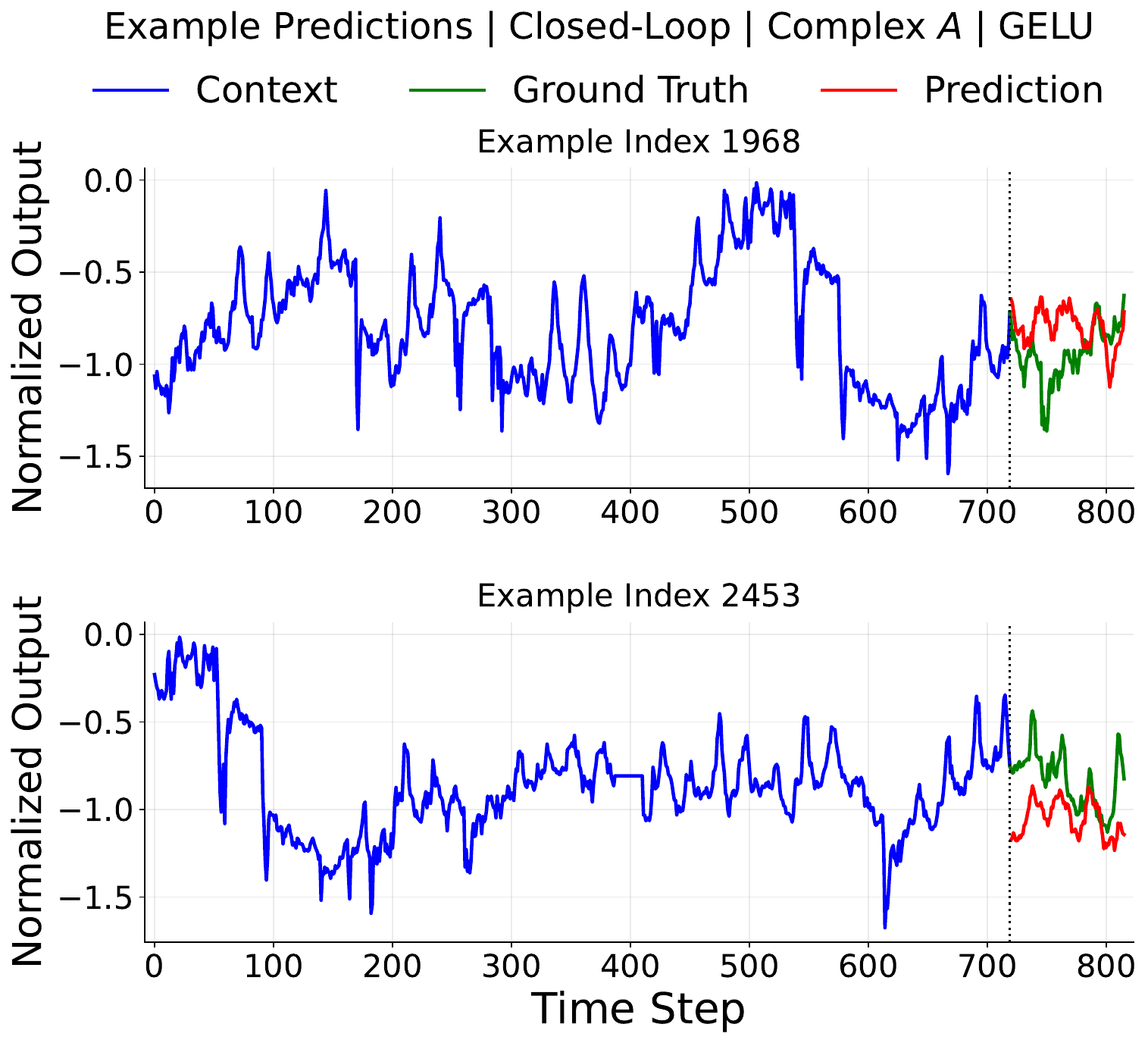}
    \caption{\SfourDCL on ETTh1.}
    \label{fig:ETT1_CL_G_0}
\end{subfigure}
\hfill
\begin{subfigure}[b]{0.32\textwidth}
    \centering
    \includegraphics[width=0.99\linewidth]{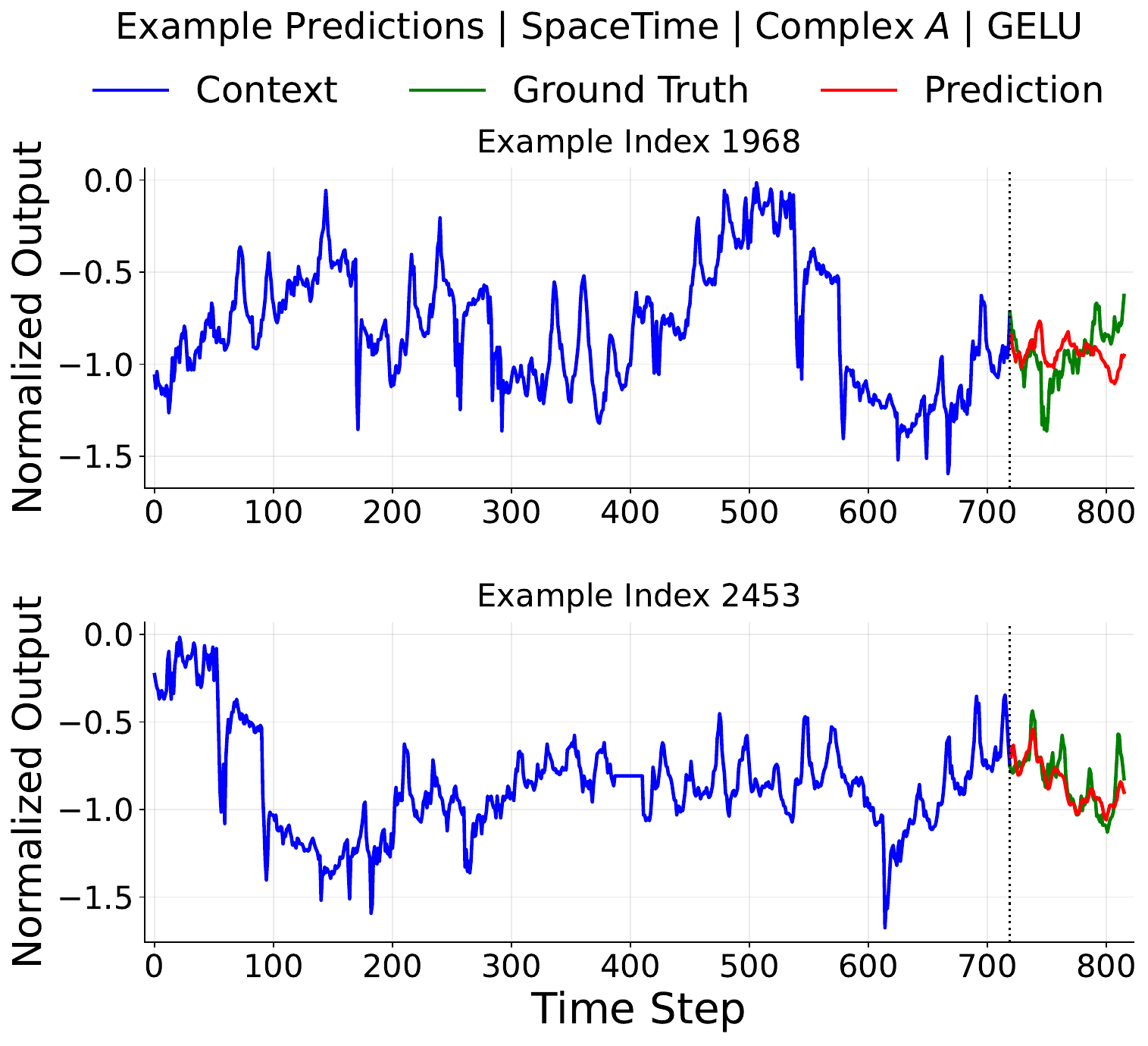}
    \caption{\SfourDST on ETTh1.}
    \label{fig:ETT1_ST_G_0}
\end{subfigure}
\vspace{0.1cm}
\begin{subfigure}[b]{0.32\textwidth}
    \centering
    \includegraphics[width=0.99\linewidth]{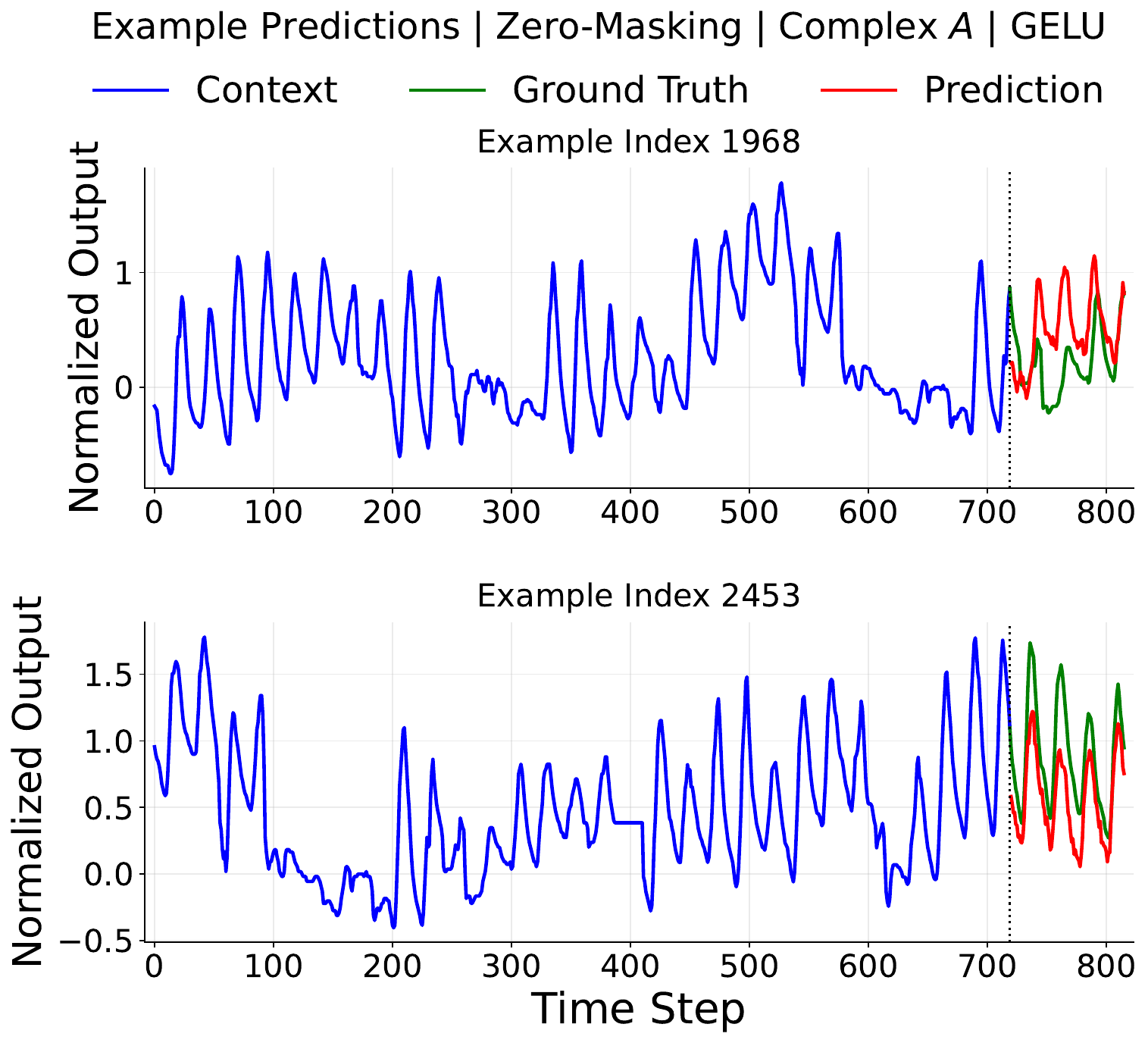}
    \caption{\SfourDZM on ETTh2.}
    \label{fig:ETT2_ZM_G_0}
\end{subfigure}
\hfill
\begin{subfigure}[b]{0.32\textwidth}
    \centering
    \includegraphics[width=0.99\linewidth]{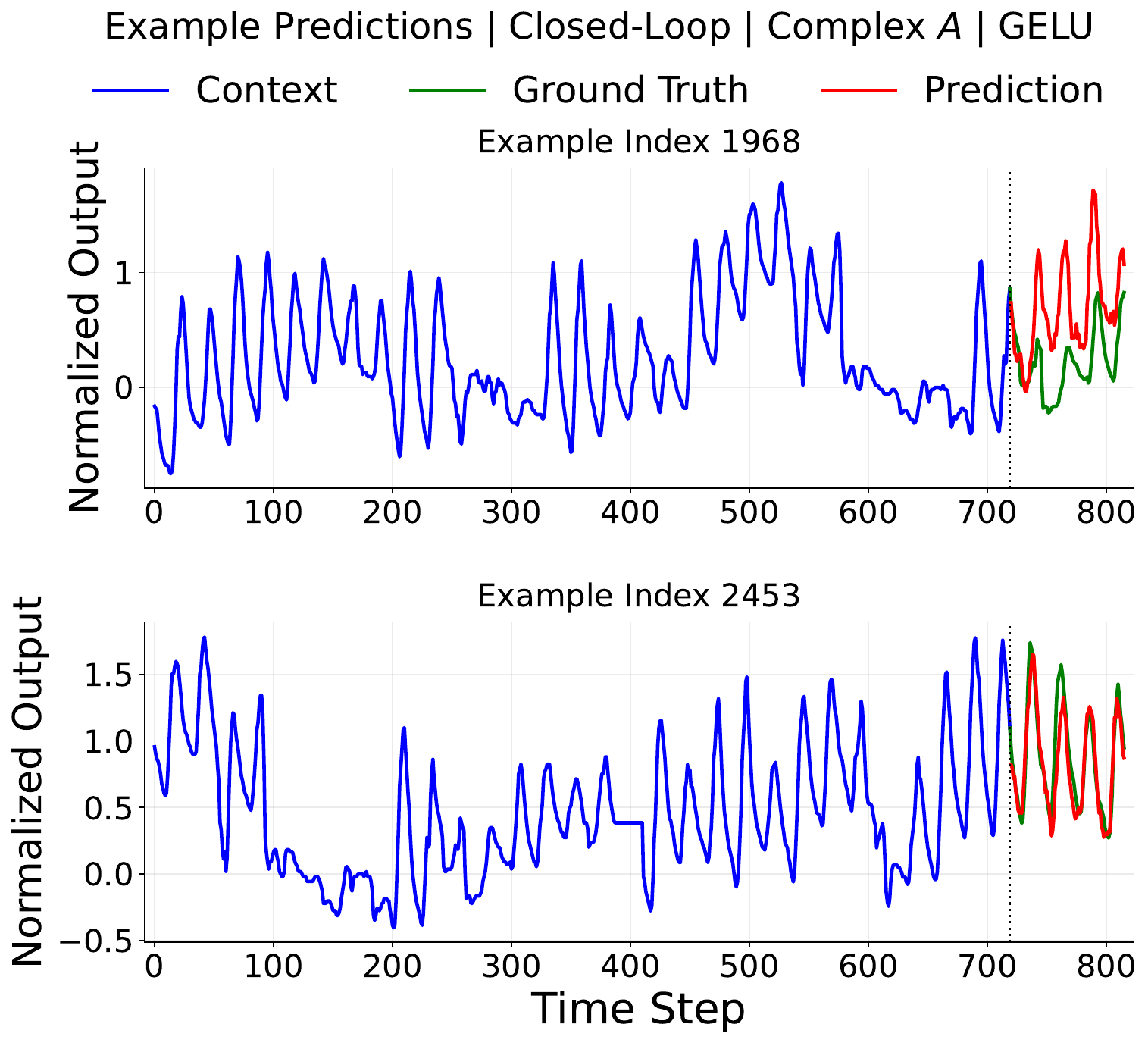}
    \caption{\SfourDCL on ETTh2.}
    \label{fig:ETT2_CL_G_0}
\end{subfigure}
\hfill
\begin{subfigure}[b]{0.32\textwidth}
    \centering
    \includegraphics[width=0.99\linewidth]{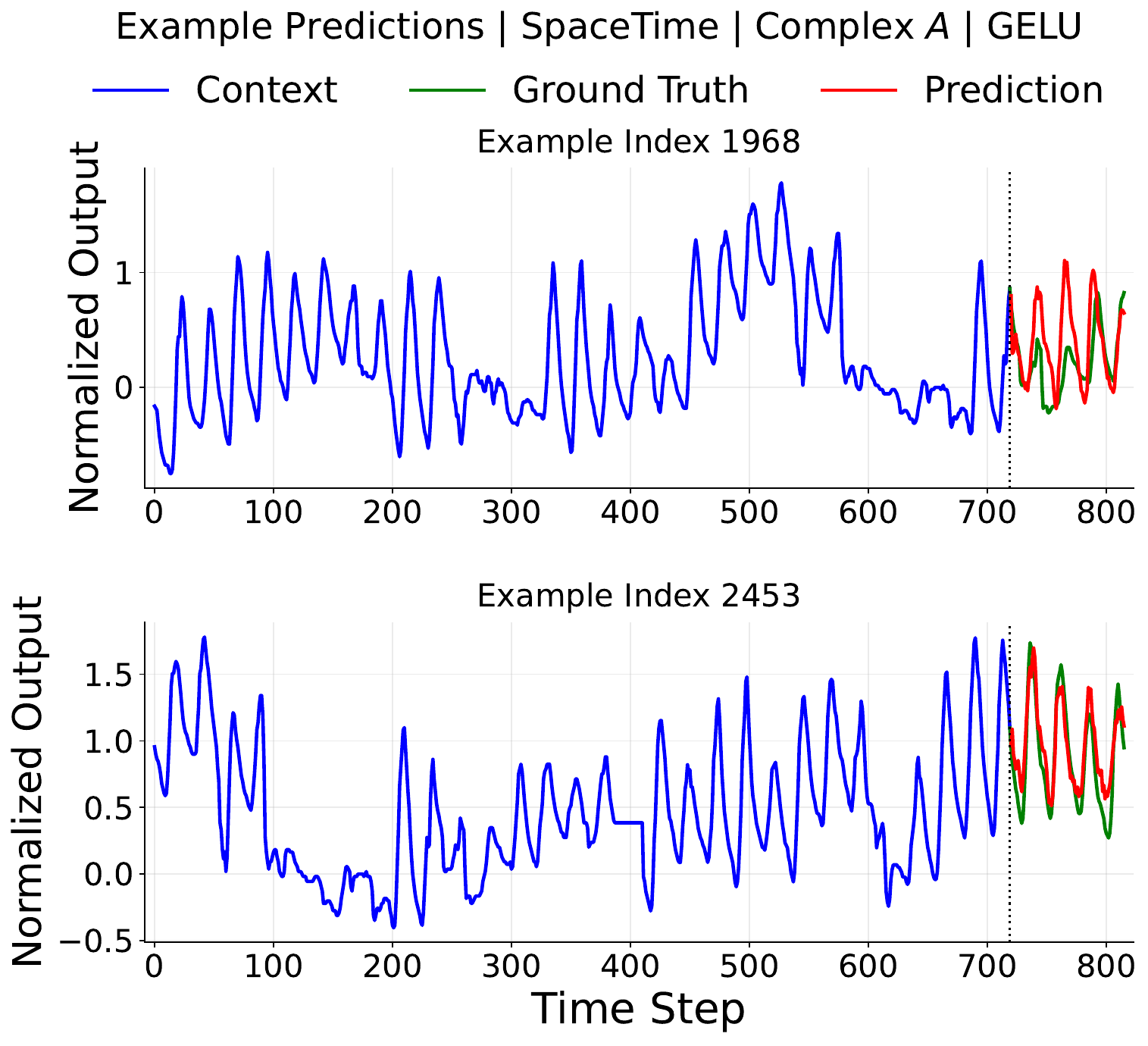}
    \caption{\SfourDST on ETTh2.}
    \label{fig:ETT2_ST_G_0}
\end{subfigure}
\vspace{0.1cm}
\begin{subfigure}{0.32\textwidth}
    \centering
    \includegraphics[width=0.99\linewidth]{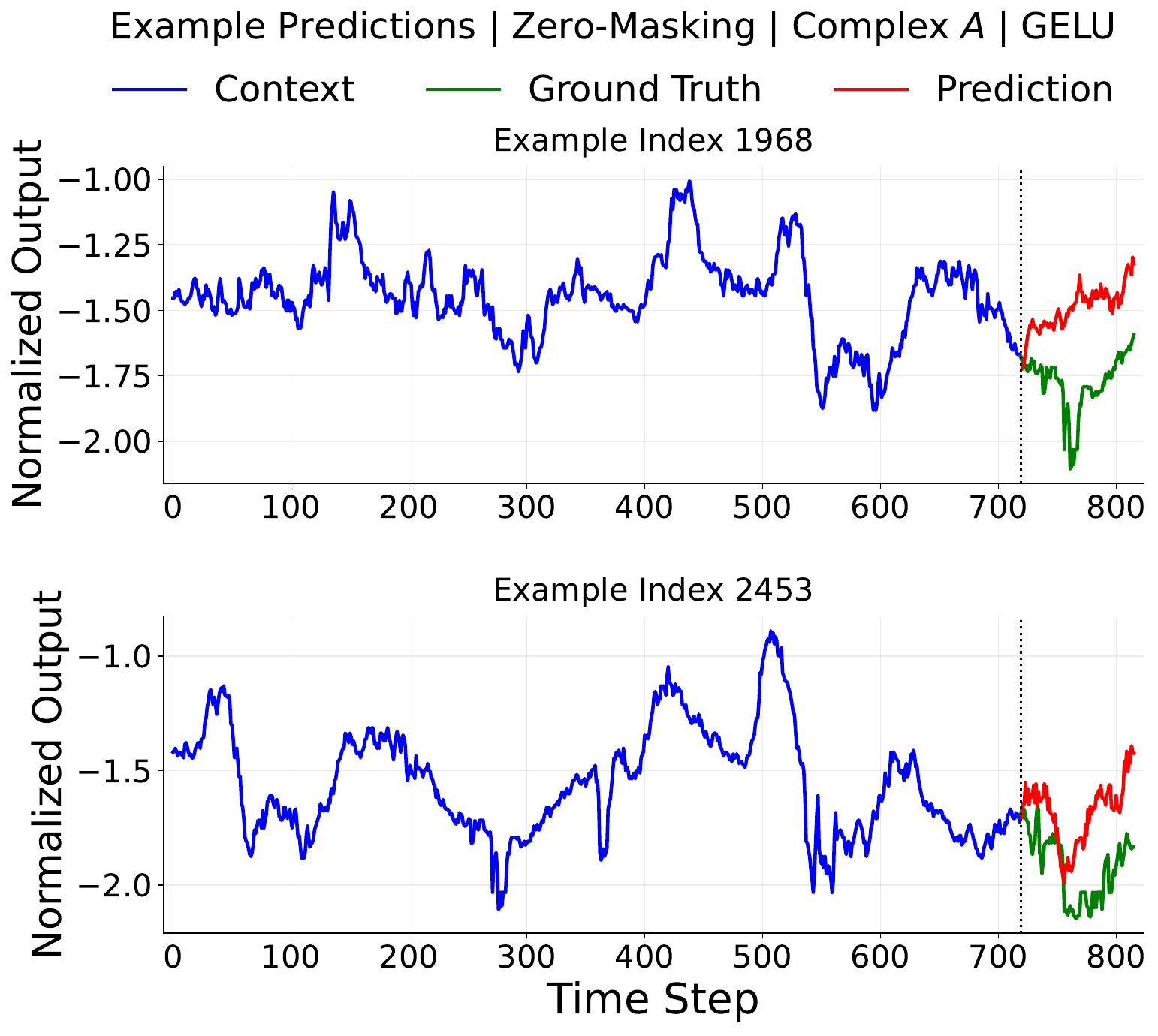}
    \caption{\SfourDZM on ETTm1.}
    \label{fig:ETT3_ZM_G_0}
\end{subfigure}
\hfill
\begin{subfigure}{0.32\textwidth}
    \centering
    \includegraphics[width=0.99\linewidth]{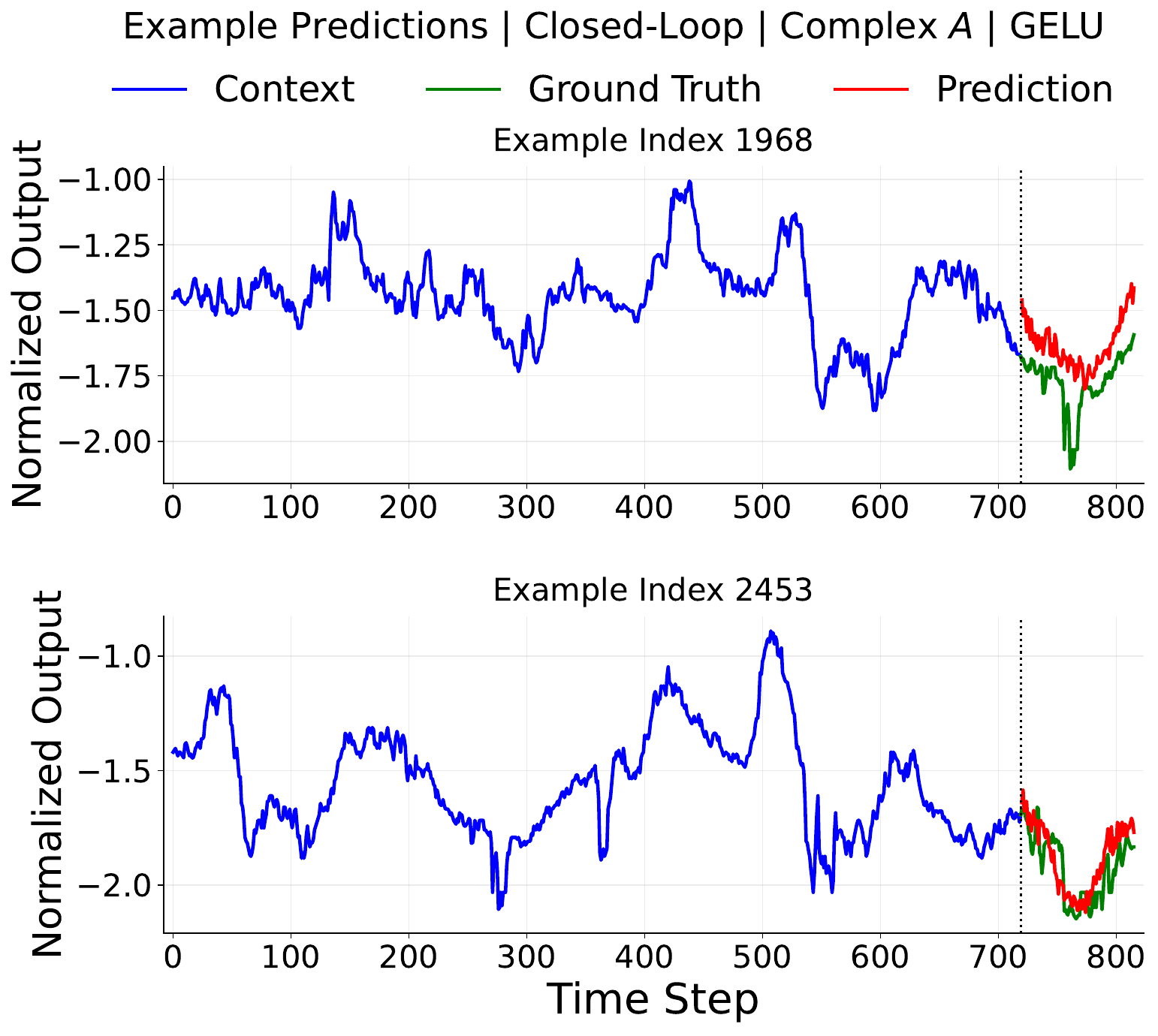}
    \caption{\SfourDCL on ETTm1.}
    \label{fig:ETT3_CL_G_0}
\end{subfigure}
\hfill
\begin{subfigure}{0.32\textwidth}
    \centering
    \includegraphics[width=0.99\linewidth]{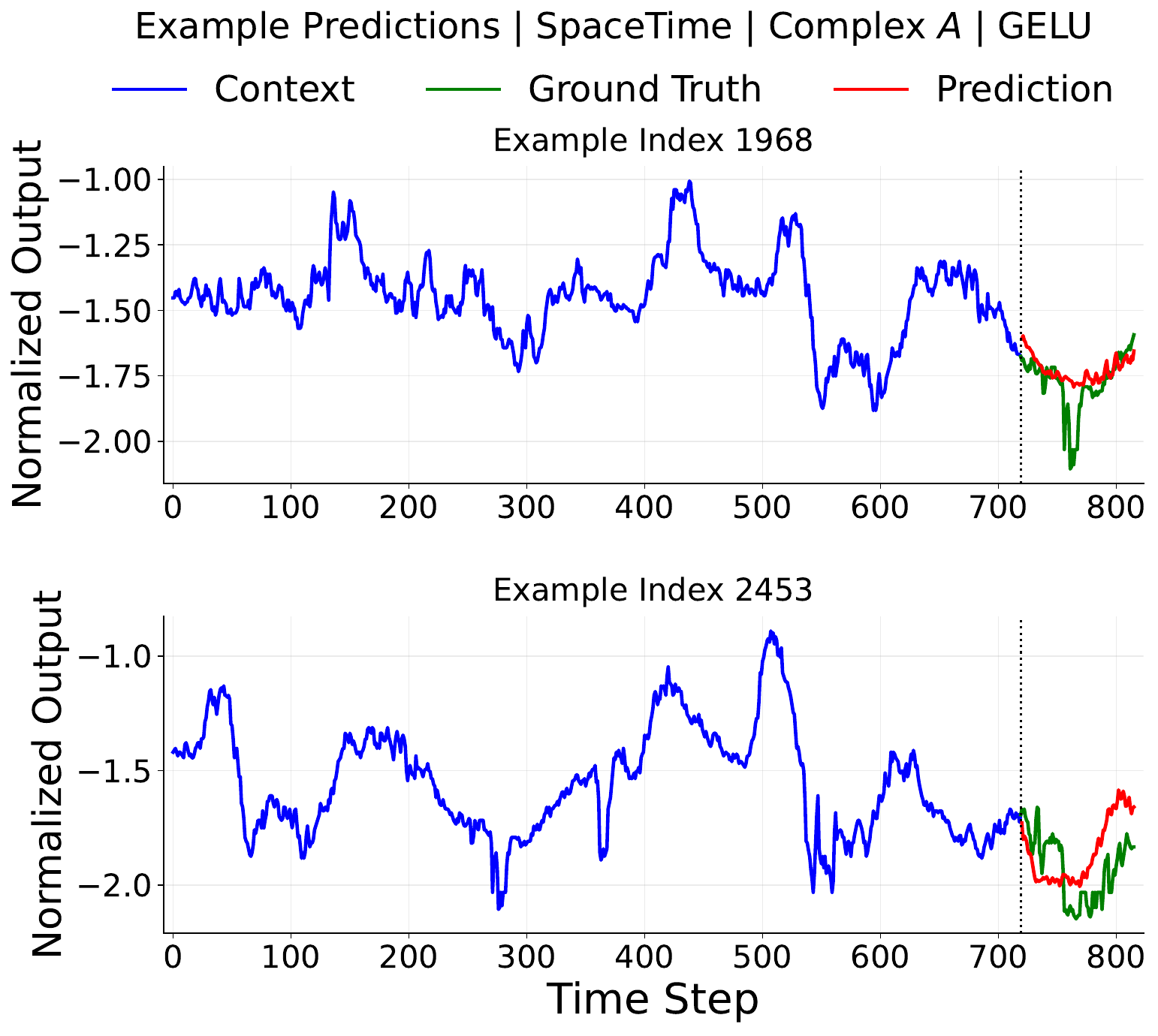}
    \caption{\SfourDST on ETTm1.}
    \label{fig:ETT3_ST_G_0}
\end{subfigure}
    \caption{Example forecasting samples for all strategies on the ETT-small datasets with context length $P=720$, forecast horizon $F=96$, and with \textbf{GELU activation}.}
    \label{fig:gelu}
\end{figure*}

Figure \ref{fig:gelu} is showing example forecasting samples for all strategies on the ETT-small datasets with GELU activation function. This is complementary to Figure 2 in the main text, which shows the same samples using Signed-Spk. activation function.

% GELU

%% file: references.bib
@book{shumway2000time,
  author    = {Shumway, Robert H. and Stoffer, David S.},
  title     = {Time Series Analysis and Its Applications},
  edition    = {3rd},
  year      = {2011},
  publisher = {Springer},
  location  = {New York},
  isbn      = {978-1-4419-7865-3},
}

@inproceedings{
gu2022efficientlymodelinglongsequences,
title={Efficiently Modeling Long Sequences with Structured State Spaces},
author={Albert Gu and Karan Goel and Christopher Re},
booktitle={International Conference on Learning Representations},
year={2022},
}

@inproceedings{zeng2022transformerseffectivetimeseries,
  author    = {Zeng, Ailing and Chen, Muxi and Zhang, Lei and Xu, Qiang},
  title     = {Are Transformers Effective for Time Series Forecasting?},
  booktitle = {Proceedings of the AAAI Conference on Artificial Intelligence},
  volume    = {37},
  number    = {9},
  pages     = {11121--11128},
  year      = {2023},
}

@misc{zhang2023effectivelymodelingtimeseries,
      title={Effectively Modeling Time Series with Simple Discrete State Spaces}, 
      author={Michael Zhang and Khaled K. Saab and Michael Poli and Tri Dao and Karan Goel and Christopher Ré},
      year={2023},
      eprint={2303.09489},
      archivePrefix={arXiv},
}

@inproceedings{haoyietal-informer-2021,
  author    = {Haoyi Zhou and
               Shanghang Zhang and
               Jieqi Peng and
               Shuai Zhang and
               Jianxin Li and
               Hui Xiong and
               Wancai Zhang},
  title     = {Informer: Beyond Efficient Transformer for Long Sequence Time-Series Forecasting},
  booktitle = {The Thirty-Fifth {AAAI} Conference on Artificial Intelligence, {AAAI} 2021, Virtual Conference},
  volume    = {35},
  number    = {12},
  pages     = {11106--11115},
  publisher = {{AAAI} Press},
  year      = {2021},
}

@inproceedings{gu2022parameterizationinitializationdiagonalstate,
  author    = {Gu, Albert and Gupta, Ankit and Goel, Karan and R{\'e}, Christopher},
  title     = {On the Parameterization and Initialization of Diagonal State Space Models},
  booktitle = {Advances in Neural Information Processing Systems (NeurIPS)},
  volume    = {35},
  year      = {2022},
}

@inproceedings{ioffe2015batchnormalizationacceleratingdeep,
  author    = {Ioffe, Sergey and Szegedy, Christian},
  title     = {Batch Normalization: Accelerating Deep Network Training by Reducing Internal Covariate Shift},
  booktitle = {Proceedings of the 32nd International Conference on Machine Learning (ICML)},
  series    = {Proceedings of Machine Learning Research},
  volume    = {37},
  pages     = {448--456},
  year      = {2015},
}

@inproceedings{lv2024efficienteffectivetimeseriesforecasting,
  author    = {Lv, Changze and Wang, Yansen and Han, Dongqi and Zheng, Xiaoqing and Huang, Xuanjing and Li, Dongsheng},
  title     = {Efficient and Effective Time-Series Forecasting with Spiking Neural Networks},
  booktitle = {Proceedings of the 41st International Conference on Machine Learning (ICML)},
  series    = {Proceedings of Machine Learning Research},
  volume    = {235},
  year      = {2024},
}

@inproceedings{shibo2025tslif,
title={{TS}-{LIF}: A Temporal Segment Spiking Neuron Network for Time Series Forecasting},
author={Fend Shibo and Wanjin Feng and Xingyu Gao and Peilin Zhao and Zhiqi Shen},
booktitle={The Thirteenth International Conference on Learning Representations},
year={2025},
}

@article{KULKARNI20133628,
title = {A spiking neural network (SNN) forecast engine for short-term electrical load forecasting},
journal = {Applied Soft Computing},
volume = {13},
number = {8},
pages = {3628-3635},
year = {2013},
issn = {1568-4946},
doi = {https://doi.org/10.1016/j.asoc.2013.04.007},
author = {Santosh Kulkarni and Sishaj P. Simon and K. Sundareswaran},
}

@article{neftci2019surrogate,
  author  = {Neftci, Emre O. and Mostafa, Hesham and Zenke, Friedemann},
  title   = {Surrogate Gradient Learning in Spiking Neural Networks: Bringing the Power of Gradient-Based Optimization to Spiking Neural Networks},
  journal = {IEEE Signal Processing Magazine},
  volume  = {36},
  number  = {6},
  pages   = {51--63},
  year    = {2019},
  doi     = {10.1109/MSP.2019.2931595},
}

@ARTICLE{10.3389/fnins.2022.865897,
    
AUTHOR={Bittar, Alexandre  and Garner, Philip N. },
           
TITLE={A surrogate gradient spiking baseline for speech command recognition},
          
JOURNAL={Frontiers in Neuroscience},
          
VOLUME={Volume 16 - 2022},
  
YEAR={2022},
  
DOI={10.3389/fnins.2022.865897},
  
ISSN={1662-453X}}

@article{davies_advancing_2021,
  author={Davies, Mike and Wild, Andreas and Orchard, Garrick and Sandamirskaya, Yulia and Guerra, Gabriel A. Fonseca and Joshi, Prasad and Plank, Philipp and Risbud, Sumedh R.},
  journal={Proceedings of the IEEE}, 
  title={Advancing Neuromorphic Computing With {Loihi}: A Survey of Results and Outlook}, 
  year={2021},
  volume={109},
  number={5},
  doi={10.1109/JPROC.2021.3067593},
}

@inproceedings{
wu2025spikf,
title={SpikF: Spiking Fourier Network for Efficient Long-term Prediction},
author={Wenjie Wu and Dexuan Huo and Hong Chen},
booktitle={Forty-second International Conference on Machine Learning},
year={2025},
}
